\documentclass[journal]{IEEEtran}

\usepackage{amssymb}
\usepackage{amsmath,amsfonts}
\usepackage[ruled,vlined,boxed,linesnumbered]{algorithm2e}
\usepackage{amsthm}
\usepackage{array}
\usepackage[caption=false,font=normalsize,labelfont=sf,textfont=sf]{subfig}
\usepackage{textcomp}
\usepackage{stfloats}
\usepackage{url}
\usepackage{multirow}
\usepackage{verbatim}
\usepackage{graphicx}
\usepackage{booktabs}
\usepackage{stackengine}
\usepackage{booktabs}
\usepackage{float}
\usepackage{graphicx}
\usepackage{stfloats}

\usepackage{hyperref}

\usepackage{amssymb}
\usepackage{amsmath}
\usepackage{amsfonts}
\usepackage{multirow}
\usepackage{graphicx}
\usepackage{textcomp}
\usepackage{amsthm}

\usepackage{setspace}
\usepackage{algorithmic}
\usepackage[ruled,vlined,boxed,linesnumbered]{algorithm2e}
\usepackage{color}
\usepackage{amsmath}
\ifCLASSINFOpdf
\else
\fi

\begin{document}

\title{FedCL: Federated Multi-Phase Curriculum Learning to Synchronously Correlate User Heterogeneity}

\author{Mingjie Wang,
    Jianxiong Guo,~\IEEEmembership{Member,~IEEE},
    and Weijia Jia,~\IEEEmembership{Fellow,~IEEE}
	\thanks{Mingjie Wang is with the Guangdong Key Lab of AI and Multi-Modal Data Processing, Department of Computer Science, BNU-HKBU United International College, Zhuhai 519087, China. (E-mail: mingjiewang@uic.edu.cn)
	
	Jianxiong Guo and Weijia Jia are with the Advanced Institute of Natural Sciences, Beijing Normal University, Zhuhai 519087, China, and also with the Guangdong Key Lab of AI and Multi-Modal Data Processing, BNU-HKBU United International College, Zhuhai 519087, China. (E-mail: jianxiongguo@bnu.edu.cn; jiawj@bnu.edu.cn)
	
	\textit{(Corresponding author: Jianxiong Guo; Weijia Jia.)}
	}
	\thanks{Manuscript received April xxxx; revised August xxxx.}}

\markboth{Journal of \LaTeX\ Class Files,~Vol.~xx, No.~xx, May~2023}%
{Shell \MakeLowercase{\textit{et al.}}: A Sample Article Using IEEEtran.cls for IEEE Journals}


\maketitle

\begin{abstract}

Federated Learning (FL) is a decentralized learning method used to train machine learning algorithms. In FL, a global model iteratively collects the parameters of local models without accessing their local data. However, a significant challenge in FL is handling the heterogeneity of local data distribution, which often results in a drifted global model that is difficult to converge. To address this issue, current methods employ different strategies such as knowledge distillation, weighted model aggregation, and multi-task learning. These approaches are referred to as asynchronous FL, as they align user models either locally or post-hoc, where model drift has already occurred or has been underestimated. In this paper, we propose an active and synchronous correlation approach to address the challenge of user heterogeneity in FL. Specifically, our approach aims to approximate FL as standard deep learning by actively and synchronously scheduling user learning pace in each round with a dynamic multi-phase curriculum. A global curriculum is formed by an auto-regressive auto-encoder that integrates all user curricula on the server. This global curriculum is then divided into multiple phases and broadcast to users to measure and align the domain-agnostic learning pace. Empirical studies demonstrate that our approach outperforms existing asynchronous approaches in terms of generalization performance, even in the presence of severe user heterogeneity.

\textit{Impact Statement}---We propose an active and synchronous correlation approach to address the challenge of user heterogeneity in FL. The proposed approach aims to approximate FL as standard deep learning by actively and synchronously scheduling user learning pace in each round with a dynamic multi-phase curriculum. It shows that the current approaches for handling user heterogeneity in FL, referred to as asynchronous FL, often result in a drifted global model that is difficult to converge. In contrast, the proposed approach aligns user models synchronously and actively by scheduling their learning pace, resulting in improved generalization performance and avoiding model drift, even in the presence of severe user heterogeneity. We employ a curriculum learning method based on loss function to integrate the learning state of users on the server, which is then divided into multiple phases and broadcast to users to measure and align the domain-agnostic learning pace. Empirical studies demonstrate that their approach outperforms existing asynchronous approaches in terms of generalization performance. Overall, the proposed active and synchronous correlation approach for handling user heterogeneity in FL offers a promising solution to the current challenges faced by FL, enabling improved convergence and generalization performance.
\end{abstract}


\IEEEpeerreviewmaketitle

\section{Introduction}
\label{sec:introduction}

\IEEEPARstart{F}{ederated} Learning (FL) enables the training of a robust deep learning model using large-scale data distributed across multiple decentralized sources without directly accessing the data. The standard FL model, represented by FedAvg \cite{mcmahan2017communication}, updates the global model by iteratively averaging the parameters of local models, thus avoiding direct contact with local user data. This privacy-preserving scheme has made FL a popular paradigm for facilitating real-world applications. To date, FL has been applied to various application areas, such as machinery fault diagnosis \cite{ZHANG2021106679}, the internet of vehicles \cite{lim2021towards}, and healthcare systems \cite{li2021federated}.

However, when applied to real-world scenarios, the standard FL model faces practical challenges arising from data heterogeneity. Local data in multiple sources or clients often have independent data sampling spaces and unique local data distributions, known as non-independent and identically distributed (Non-IID) data. This inconsistency between local objective functions and global optimization directions frequently leads to problems. Studies in \cite{khaled2020tighter}\cite{lee2021preservation}\cite{hsu2019measuring} have demonstrated that applying the standard FL model to Non-IID data, such as with FedAvg, results in drifting local models and catastrophic forgetting of global information. Consequently, this approach leads to suboptimal convergence speed and model performance. Thus, addressing Non-IID local datasets has become a crucial topic in FL.

From the above analysis, it is clear that the standard FL model with element-wise averaging operations, such as FedAvg, is not ideal for real-world applications. This has attracted numerous researchers to focus on addressing heterogeneity in FL. Current methods can be broadly categorized into four groups as follows:

\begin{itemize}
\item \textbf{Alignment of Global and Local Objectives \cite{Gao_2022_CVPR}\cite{yang2022personalized}:} This approach aims to mitigate model drift by modifying the local target to align with a desired global target, thus harmonizing global and local objectives. However, this approach's limitation is that it can only be applied to clients due to inherent structural differences in their partial local data.

\item \textbf{Weighted Aggregation \cite{9847055}\cite{9442814}\cite{li2020federated}\cite{shi2021fed}:} This method achieves aggregation by assigning weights to local models, optimizing fusion performance for the global model at the server. Nevertheless, the drawback of this approach is its reliance on proxy data, and it occurs after model drift has already taken place.

\item \textbf{Knowledge Distillation \cite{9879661}\cite{jiang2020federated}\cite{zhu2021data}:} This technique combines real data with synthesized data, aiming for a more balanced data distribution at the client or server level to align multi-client domain knowledge. However, this approach's shortcoming is that the generator model, which produces synthetic data, is also trained on Non-IID data. Consequently, the generator model is affected by the inherent data heterogeneity and model drift.

\item \textbf{Multi-task Learning \cite{9763764}\cite{he2021spreadgnn}:} This method enables federated personalization by learning a generic and personalized FL at clients. However, the disadvantage of this approach is that multi-task learning and personalized FL inherently contradict each other.

\end{itemize}

\begin{figure}[!t]
\includegraphics[width=\linewidth]{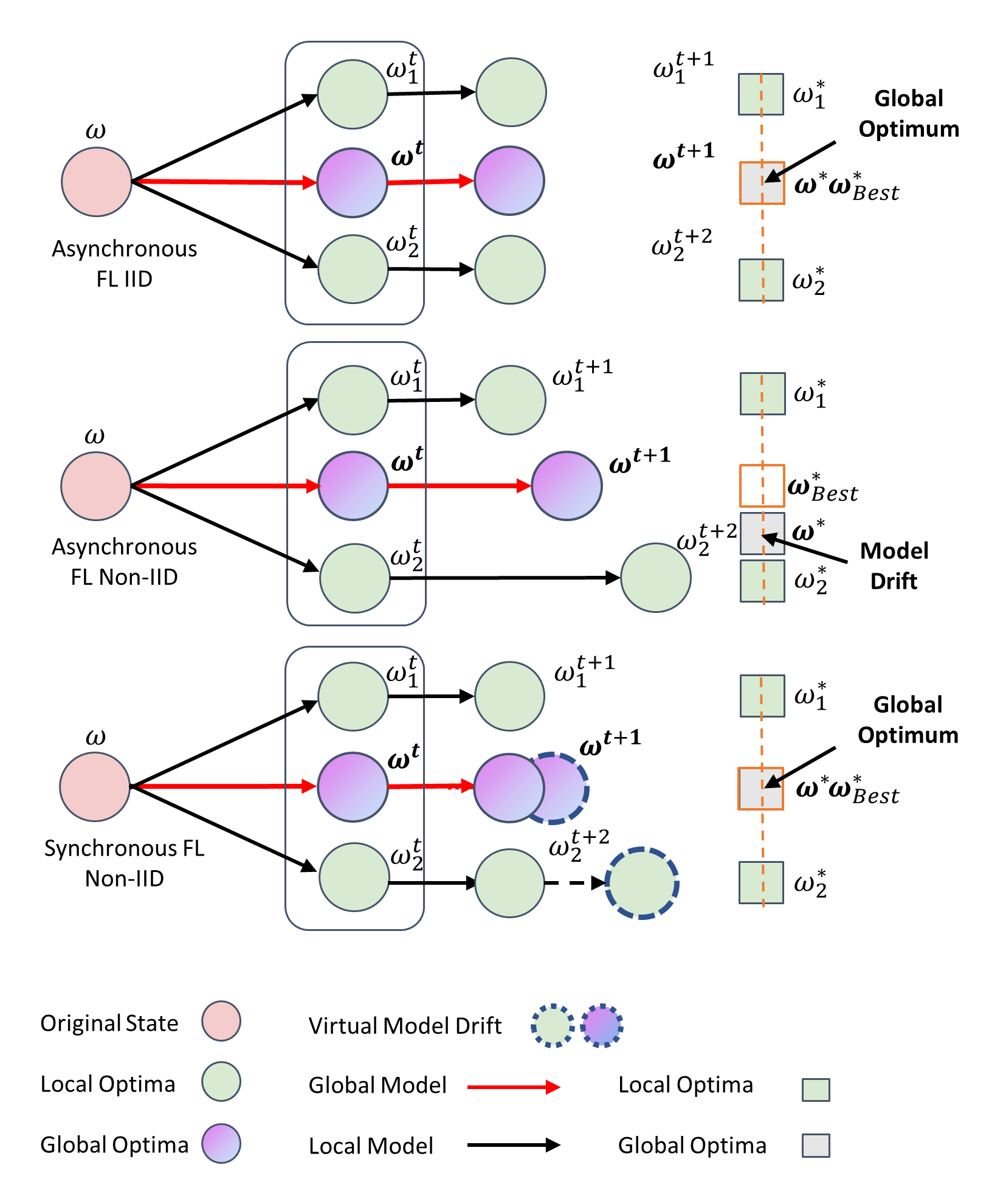}
\caption{Illustration of asynchronous and synchronous FL model under IID and Non-IID datasets, where $\boldsymbol{\omega}_1^t$ and $\boldsymbol{\omega}_2^t$ are training parameters of two local m odel at time $t$, $\boldsymbol{\omega}^t$ is training parameter of global model, $\boldsymbol{\omega}^*$ is the optimum obtained by this model, and $\boldsymbol{\omega}^*_{best}$ is the optimum under IID dataset.}
\label{IID}
\end{figure}

We observe that these approaches either underestimate heterogeneity based on inconsistent principles (client drift mitigation and federated personalization) or address model drift in a post-hoc manner (knowledge distillation and weighted model aggregation). We refer to them as asynchronous FL, considering the fitting degree of the model simultaneously. However, existing post-hoc calibration methods produce overly confident predictions under domain shift, and the prerequisite of proxy data can leave such an approach susceptible to overfitting the model.

Given the challenges of user heterogeneity and the limitations of previous asynchronous FL approaches, we adopt a synchronous FL model by approximating the learning speed of the local model to that of the centralized deep learning model, leveraging consistency and online heterogeneity alignment. 

Specifically, model aggregation under IID data consistently and efficiently converges to a relative global optimum, as shown in Figure \ref{IID} (asynchronous IID) \cite{pmlr-v108-bayoumi20a}. In contrast, Figure \ref{IID} (asynchronous Non-IID) depicts the global model $\boldsymbol{\omega}^*$ obtained from the model aggregation at the server, trained by Non-IID data in clients. This process disregards local knowledge incompatibility, leading to knowledge forgetting and performance degradation in the global model \cite{NEURIPS2020_fb269786}\cite{lee2021preservation}.

Curriculum learning (CL) is a technique that aims to enhance the generalization ability of models by progressively increasing the difficulty level of the training data. This approach has been demonstrated to improve model performance across a variety of tasks. However, in distributed learning scenarios, such as federated learning (FL), additional challenges need to be addressed, including ensuring consistency and reliability of local training and global aggregation. In this article, we propose a method called Federal Curriculum Learning (FedCL), which is based on synchronous learning and aims to overcome these challenges. FedCL comprises three components: local CL, global curriculum ensemble, and multi-phase correlation. Firstly, each client in the FL system learns a local curriculum to evaluate model-aware difficulty based on confidence. This enables clients to adjust their learning strategies according to individual progress. Secondly, the global server summarizes the clients' progress to measure performance change based on global learning progress, which is then broadcast to all users. This ensures that all users benefit from globally consistent learning progress that reflects online synchronization across all users. Lastly, based on this global learning progress, we further align the learning progress from the global model to users' local networks incrementally by incorporating a self-paced learning methodology. Our experiments demonstrate that FedCL outperforms existing FL methods on several benchmark datasets, including CIFAR-10, CIFAR-100, and MNIST. The results indicate that FedCL exhibits better generalization ability and can effectively mitigate the negative impact of heterogeneous data distribution in FL systems. Additionally, we analyze the influence of different hyperparameters on FedCL's performance and provide guidelines for selecting appropriate hyperparameters. In summary, our main contributions can be summarized as follows:

\begin{itemize}
    \item In contrast to previous asynchronous learning approaches, we propose a synchronous FL method that aggregates local models before they drift, effectively avoiding model drift and enabling clients to achieve competitive global model performance.
    
    \item Our method directly synchronizes each local model, arranging a hierarchical curriculum for personalized FL based on efficient progress synchronization and online heterogeneity alignment. Experiments demonstrate that this curriculum reduces the risk of inductive bias and overfitting of local models, leading to better generalization on Non-IID data.
    
    \item Comprehensive empirical studies, supported by theoretical explanations, show that FedCL achieves superior generalization performance using fewer communication rounds. Furthermore, we conduct thorough ablation studies, which corroborate the advantages of FedCL.
    
    \item We perform extensive experiments to validate the effectiveness of our approach in various environments, demonstrating that FedCL exhibits better generalization using fewer communication rounds.
\end{itemize}

\textbf{Organization: }Section \uppercase\expandafter{\romannumeral2} summarizes the works related this paper. Section \uppercase\expandafter{\romannumeral3} establishes the basic settings of our FL algorithm and introduces that other important approaches. Section \uppercase\expandafter{\romannumeral4} expands our model algorithm in detail which establishes a multi-phase CL approach to deal with user heterogeneity. Besides, Section \uppercase\expandafter{\romannumeral5} sets up the experiments which compare with other baselines on different datasets and parameters, and then discussions the results. Finally, Section \uppercase\expandafter{\romannumeral6} summarizes our research and future works.

\section{Related work}
\label{sec:Related work}
In this paper, we discuss Federated Learning (FL), a distributed machine learning framework initially proposed by Google. FL enables multiple clients to collaborate in training a model without disclosing their data. FedAvg \cite{mcmahan2017communication} is a representative FL method, where each client trains the model locally and uploads its parameters to the server. The server then updates the global model by averaging the local model parameters and returns it to all clients for the next training round. However, FedAvg's performance significantly declines in the case of Non-IID data, where data imbalance and differing local client distributions lead to global model divergence. This divergence limits the model's relevance and accuracy, resulting in increased communication rounds required for convergence.

\subsubsection{Federated Learning with Heterogeneous Data}
Several studies aim to improve FedAvg on heterogeneous data. These studies can be divided into four categories: alignment of global and local objectives, weighted aggregation, knowledge distillation, and multi-task learning. We will introduce them individually here.

In terms of research on improving the consistency of global and local objectives, FedProx \cite{li2020federated} enhances the local training model by incorporating an approximation term into the target during the local training process. It can train and predict various natural distributions with good performance on different natural distribution offsets owned by various clients. FedDC \cite{Gao_2022_CVPR} employs learned auxiliary local drift variables to bridge the gap between local and global model parameters, achieving parameter-level consistency. However, differences in local data information structure limit each user's effectiveness in learning the global goal and the aggregation model's consistency. Regarding studies on improving weighted aggregation, Fedensemble \cite{shi2021fed} uses random arrangement to update a set of $k$ models, averaging these models to obtain predictions. FedAdp \cite{9442814} adaptively assigns different weights to update the global model according to node contributions. Each user's weight is calculated based on the similarity between the gradients of all local models, restricting local model updating due to the risk of model drift during local training. In knowledge extraction research, FedDistill \cite{jiang2020federated} extracts anonymous data from user model logit-vectors and shares this meta-data with users for knowledge distillation. FedGen \cite{zhu2021data} proposes a generator to produce data to balance the distribution of corresponding features under the given prediction, reducing the influence of data heterogeneity. The effectiveness of knowledge extraction is limited by local model and generator performance, and there remains the risk of local model drift. For multi-task learning research, CFMTL \cite{9763764} uses a clustering method to measure local data similarity among groups, allowing clients with similar data distribution to be grouped, transitioning from single tasks to group tasks. Regrettably, multi-task learning constrains the upper limit of personalization and global optimization of local models. Moreover, under extreme Non-IID data, multi-task learning will degenerate into the primitive FedAvg.

The existing methods primarily employ asynchronous FL, which modifies partial or posterior user models or specific aggregated drift models, leading to over-fitted or under-fitted models. Consequently, we propose a synchronous FL method based on Curriculum Learning, aligning the global model with local models in terms of model fitting. Curriculum Learning can assess each local model's learning situation and the global model's learning progress, ensuring personalization and consistency of FL under Non-IID data.

\subsubsection{Curriculum Learning}

Curriculum Learning (CL) is an approach that assigns different weights to samples, guiding the model learning process to follow a sequence from easy to difficult. CL can accelerate the training of machine learning models, reduce the number of iterations, and achieve a better local optimal state \cite{bengio2009curriculum}. Most CL designs are currently based on the framework of difficulty measurement and training scheduler, with Automatic CL offering more flexibility since it can automatically participate and consider its feedback during the process. Automatic CL improvements can be divided into three categories: optimizing the selection of learning samples, addressing dimension and edge problems through self-supervised learning, and optimizing local personalization.

In research on optimizing the selection of learning samples, the training sample of CL transitioned from relying on human experts to avoid local minima \cite{Basu_Christensen_2013} to automatically selecting strategies using various standard samples during training \cite{graves2017automated}. Hacohen \textit{et al.} \cite{hacohen2019power} show that, under mild conditions, it does not change the corresponding global minimum of the optimization function. The selection criteria of these methods were developed for various learning settings and hypothesis class assumptions, revealing that many CL methods prefer easy-to-learn samples \cite{jiang2015self}\cite{kumar2010self}.

Regarding optimizing dimension and edge problems, researchers have considered self-supervised learning as a means to address these issues. Specifically, Kumar \textit{et al.} \cite{kumar2010self} proposed a method that considers examples with a loss, while Katharopoulos \textit{et al.} \cite{katharopoulos2018not} correlated the importance of a sample with its norm of loss gradient of network parameters. In the context of noisy examples, Jiang \textit{et al.} \cite{jiang2018mentornet} examined the robustness of CL and proposed Curriculum Accelerated Self-Supervised Learning (CASSL), which identifies memorable examples that can be safely deleted without affecting generalization and demonstrates that noisy examples rank higher in terms of the number of forgotten events. Additionally, to address the edge problem of CL optimization, a covering model has been proposed that achieves better generalization and improved convergence rates \cite{basu2013teaching}.

In the realm of optimizing local personalization, to address the problem of heterogeneous data in virtual reality, many studies have been devoted to training the best personalized models for each customer \cite{sattler2020clustered}. A general curriculum learning strategy (DoCL) was developed for the optimization of training dynamics \cite{zhou2021curriculum}. Yang \textit{et al.} \cite{yang2022hybrid} propose a model-agnostic hybrid CL strategy.

In our work, CL is utilized to alleviate the heterogeneity of local data and unify client learning progress. Due to the heterogeneity of data, the difficulty lies in helping the model approximate the difficulty distribution of generated data to that of global data. Therefore, we have added some concepts to guide the optimization process through curriculum learning, so as to better accomplish the final task in the heterogeneous environment.

\section{NOTATIONS \& PRELIMINARIES}

This section presents the mathematical formulation and objective function in FL. After introducing the concept of CL, we redefine the FL objective function used in this paper. Furthermore, to ensure data privacy during client-to-server transmission, the Gaussian Mixture Model is introduced.

\subsection{Problem Notations}
Federated Learning (FL) is a machine learning paradigm that aims to protect user privacy by allowing participants to collaborate on modeling without sharing data. Consider a typical FL setting for supervised learning of a multi-class classification problem, we define the input feature space as $\mathcal{X}\subset \mathbb{R}^{p}$, the latent feature space as $\mathcal{Z} \subset \mathbb{R}^{d}$ with $d<p$, and the output space as $\mathcal{Y} \subset \mathbb{R}$. $\mathcal{T}$ is a domain (or called task), denoted by $\mathcal{T}:=\left\langle\mathcal{D}, c^{*}\right\rangle$, which has a data distribution $\mathcal{D}$ over $\mathcal{X}$ and a ground-truth labeling function $c^*: \mathcal{X}\rightarrow\mathcal{Y}$. A model can be parameterized by $\boldsymbol{\omega}:=\left[\boldsymbol{\omega}^{f} ; \boldsymbol{\omega}^{h}\right]$, which consists of two components: a feature extractor $f: \mathcal{X} \rightarrow \mathcal{Z}$ parameterized by $\boldsymbol{\omega}^{f}$ and a predictor $h: \mathcal{Z} \rightarrow\Delta^{\mathcal{Y}}$ parameterized by $\boldsymbol{\omega}^{h}$, where $\Delta^{\mathcal{Y}}$ is the simplex over $\mathcal{Y}$. By giving a non-negative and convex loss function $l:\Delta^{\mathcal{Y}}\times\mathcal{Y} \rightarrow \mathbb{R}$, the expected risk of this model parameterized by $\boldsymbol{\omega}$ on domain $\mathcal{T}$ can be defined as follows. That is:
\begin{equation}
\mathcal{L}_{\mathcal{T}}(\boldsymbol{\omega}):= \mathbb{E}_{x \sim \mathcal{D}}\left[l\left(h\left(f\left(x ; \boldsymbol{\boldsymbol{\omega}}^f\right) ; \boldsymbol{\boldsymbol{\omega}}^h\right), c^*(x)\right)\right],
\end{equation}

\subsection{Federated Learning}
\label{fl}
In FL, the optimization objective is represented by $\mathcal{L}$ for the global task $\mathcal{T}$ and by $\mathcal{L}_{k}$ for the local task $\mathcal{T}_k$. The aim of FL is to learn a global model, parameterized by $\boldsymbol{\omega}$, that minimizes the risk on each client task. The global objective function is expressed as follows \cite{mcmahan2017communication}:
\begin{equation}
\label{eq1}
\min\limits_{\boldsymbol{\omega}}\mathbb{E}_{\mathcal{T}_k \in \mathcal{T}}\left[\mathcal{L}_{k}(\boldsymbol{\omega})\right] ,
\end{equation}
where $\mathcal{T}_k =\left\langle\mathcal{D}_k, c^{*}\right\rangle$. Assuming there are total $K$ clients, denoted by $[K]$. In detail, for each client $k\in[K]$, its local task is $\mathcal{T}_k =\left\langle\mathcal{D}_k, c^{*}\right\rangle$, and we have:
\begin{equation}
\label{localloss}
\mathcal{L}_{k}(\boldsymbol{\omega}):= \mathbb{E}_{x\sim \mathcal{D}_k}\left[l\left(h\left(f\left(x ; \boldsymbol{\boldsymbol{\omega}}^f\right) ; \boldsymbol{\boldsymbol{\omega}}^h\right), c^*(x)\right)\right],
\end{equation}
where $\mathcal{T}=\left\{\mathcal{T}_k\right\}_{k=1}^K$ is the collection of client tasks $\mathcal{T}_k=\left\langle\mathcal{D}_k, c^*\right\rangle$ for each client $k\in[K]$ share the same mapping rules $c^*$ and loss function $l$. In practice, Eqn. (\ref{eq1}) can be empirically optimized by
\begin{equation}
\min\limits_{\boldsymbol{\omega}}\frac{1}{K} \sum\nolimits_{k=1}^{K} \hat{\mathcal{L}}_{k}(\boldsymbol{\omega}),
\label{globalobj}
\end{equation}
where we have $\hat{\mathcal{L}}_k(\boldsymbol{\omega})$ , whcih is the empirical risk over an an observable dataset $\hat{\mathcal{D}}_k$:
\begin{equation}\label{eq5}
    \hat{\mathcal{L}}_k(\boldsymbol{\omega}):=\frac{1}{|\hat{\mathcal{D}}_k|} \sum_{x_i \in \hat{\mathcal{D}}_k}\left[l\left(h\left(f\left(x_i ; \boldsymbol{\omega}^f\right) ; \boldsymbol{\omega}^h\right), c^*\left(x_i\right)\right)\right],
\end{equation}
Thus, in this FL model, the global dataset $\hat{\mathcal{D}}$ is distributed to each of the local domains, thus we have $\hat{\mathcal{D}}=\cup\{\hat{\mathcal{D}}_k\}_{k=1}^K$.

\subsection{Curriculum Learning}
Curriculum Learning (CL) is a method that trains a model by assigning different weights to samples based on the difficulty of learning each sample. We introduce a CL loss function inspired by confidence-aware \cite{NEURIPS2020_2cfa8f9e}, which is simply integrated with the original task loss during training to monitor each sample's loss and dynamically determine sample contributions by applying the core principle of curriculum learning. The confidence-aware loss function ${l}^{cl}(\cdot)$ consists of a difficulty score loss function $\bar{l}^{cl}(\cdot)$ and a regularization term. The difficulty score follows the task core of CL and multiplies the confidence parameter in a form of the loss function to achieve the weight distribution from easy to difficult. Let $x_{i}$ be the $i^{th}$ sample of dataset $\hat{\mathcal{D}}_k$ in the $k^{th}$ client. The difficulty score is calculated as follows:

\begin{equation}\label{diff}
\centering
\begin{aligned}
\bar{l}^{cl}&\left(h\left(f\left(x_{i};\boldsymbol{\omega}^{f}\right);\boldsymbol{{\omega}}{^{h}}\right), c^{*}(x_{i})\right)\\
&=\left[l\left(h\left(f\left(x_{i};\boldsymbol{\omega}^{f}\right);\boldsymbol{{\omega}}{^{h}}\right), c^{*}(x_{i})\right)-\tau\right]\cdot\sigma_{i},
\end{aligned}
\end{equation}
Where $\sigma_{i}$ is a confidence parameter and $\tau$ represents a loss-based threshold distinguishing easy and hard samples, we aim to avoid overfitting issues resulting from high model complexity. We add a regularization term $\lambda\left(\log \sigma_{i}\right)^{2}$ to Eqn. \eqref{diff} to minimize structural risk, where $\lambda$ is the hyper-parameter. Consequently, we can formulate the confidence-aware loss function as follows:
\begin{equation}
\label{lcl}
\centering
\begin{aligned}
l^{cl}&\left(h\left(f\left(x_{i} ; \boldsymbol{\omega}^{f}\right);\boldsymbol{{\omega}}{^{h}}\right), c^{*}(x_{i})\right)\\
&=\bar{l}^{cl}\left(h\left(f\left(x_{i};\boldsymbol{\omega}^{f}\right);\boldsymbol{{\omega}}{^{h}}\right), c^{*}(x_{i})\right)+\lambda\left(\log \sigma_{i}\right)^{2},
\end{aligned}
\end{equation}
By using $l^{cl}$ in Eq. \eqref{lcl} to redefine local object ${\mathcal{L}}_{k}$ in Eq. \eqref{localloss} as:
\begin{equation}
\mathcal{L}_k^{cl}(\boldsymbol{\omega}):=\mathbb{E}_{x \sim \mathcal{D}_k}\left[l^{cl}\left(h\left(f\left(x_i ; \omega^f\right) ; \omega^h\right), c^*\left(x_i\right)\right)\right],
\label{newlocal}
\end{equation}
A change of local goals also needs to redefine global goals,
the redefined global object function in Eq. \eqref{eq1} as:
\begin{equation}
\min _{\boldsymbol{\omega}} \mathbb{E}_{\mathcal{T}_k \in \mathcal{T}}\left[\mathcal{L}_k^{cl},(\boldsymbol{\omega})\right]
\end{equation}
Similarly, we use the confidence-aware loss $l^{cl}$ instead of $l$ in the empirical risk $\hat{\mathcal{L}}_{k}$ shown as Eqn. \eqref{eq5}. Then, the $\hat{\mathcal{L}}_{k}$ can be redefined by $\hat{\mathcal{L}}_{k}^{cl}$ as:
\begin{equation}
\hat{\mathcal{L}}_{k}^{cl}(\boldsymbol{\omega})=
\frac{1}{|\hat{\mathcal{D}}_k|} \sum_{x_i \in \hat{\mathcal{D}}_k} l^{cl}\left(h\left(f\left(x_{i} ; \boldsymbol{\omega}^{f}\right);\boldsymbol{{\omega}}{^{h}}\right), c^{*}(x_{i})\right).
\label{local}
\end{equation}
Back to the task in Section. \ref{fl}, FL finally expresses a standard way of defining a global loss function in the context of optimization. The goal is to find the value of $\boldsymbol{\omega}$ that minimizes the average loss across all data points as:
\begin{equation}\label{end}
    \min\limits_\omega \frac{1}{K} \sum\nolimits_{k-1}^K \hat{\mathcal{L}}_{k}^{cl}(\boldsymbol{\omega}),
\end{equation}
where it redefines the global loss function in Eqn. \eqref{globalobj} by the sum over $k$ in the expression.

\subsection{Gaussian Mixture Model}
\label{GMM}
The Gaussian Mixture Model (GMM) is a probability model positing that all data points are generated by a finite number of unknown parameters, forming a mixture of Gaussian distributions. Theoretically, any continuous distribution can be approximated using a finite number of Gaussian distribution mixtures. Each client $k$ has $|\hat{\mathcal{D}}k|$ samples, and for each observation sample $x_i\in\hat{\mathcal{D}}_k$, we can obtain its difficulty score $\bar{l}^{c l}(h(f(x{_i} ; \boldsymbol{\omega}^f) ; \boldsymbol{\omega}^h), c^*(x{_i}))$ as shown in Eqn. \eqref{diff}. The distribution of difficulty scores over all samples in client $k$ can be denoted by $\mathcal{D}^{cl}_k$. We assume that the local difficulty score can be expressed as the superposition of a series of Gaussian distributions.

For each client $k$, we can estimate the distribution ${\mathcal{D}}^{cl}_{k}$ by using the difficulty score $\bar{l}^{cl}(x_i)$ for each sample $x_i\in\hat{\mathcal{D}}_k$ where we denote by $\bar{l}^{cl}(x_i)=\bar{l}^{c l}(h(f(x_{ i} ; \boldsymbol{\omega}^f) ; \omega^h), c^*(x_{ i}))$ for convenience. Given a fixed positive integer $L$, the distribution of $\mathcal{D}^{cl}_k$ can be formulated as:
\begin{equation}\label{localdis}
\mathcal{D}^{cl}_k(x;\boldsymbol{\mu}_k,\boldsymbol{\sigma}^2_k)=\sum_{l=1}^L\alpha_{k,l}\cdot\phi(x;\mu_{k,l},\sigma^2_{k,l}),
\end{equation}
where $\phi(x;\mu_{k,l},\sigma^2_{k,l})$ is a Gaussian distribution with the mean $\mu_{k,l}$ and variance $\sigma^2_{k,l}$, $\alpha_{k,l}$ is the weight of the $l$-th Gaussian distribution with $\sum_{l=1}^L\alpha_{k,l}=1$, and we have $\boldsymbol{\mu}_k=\{\mu_{k,1},\cdots,\mu_{k,N}\}$ and $\boldsymbol{\sigma}^2_k=\{\sigma^2_{k,1},\cdots,\sigma^2_{k,N}\}$. Here, we denote by $\boldsymbol{\alpha}_k=\{\alpha_{k,1},\cdots,\alpha_{k,N}\}$, and then the parameters of the distribution $\mathcal{D}^{cl}_k$ can be defined as $(\boldsymbol{\mu}_k,\boldsymbol{\sigma}^2_k,\boldsymbol{\alpha}_k)$. In order to ensure the privacy of clients while enabling the central server to obtain the global difficulty score, the parameters $(\boldsymbol{\mu}_k,\boldsymbol{\sigma}^2_k,\boldsymbol{\alpha}_k)$ should be transmitted to the server instead of the observable dataset $\hat{\mathcal{D}}_k$ from the client $k$.

\section{METHODOLOG}
\begin{figure}[!t]
\includegraphics[width=0.5\textwidth]{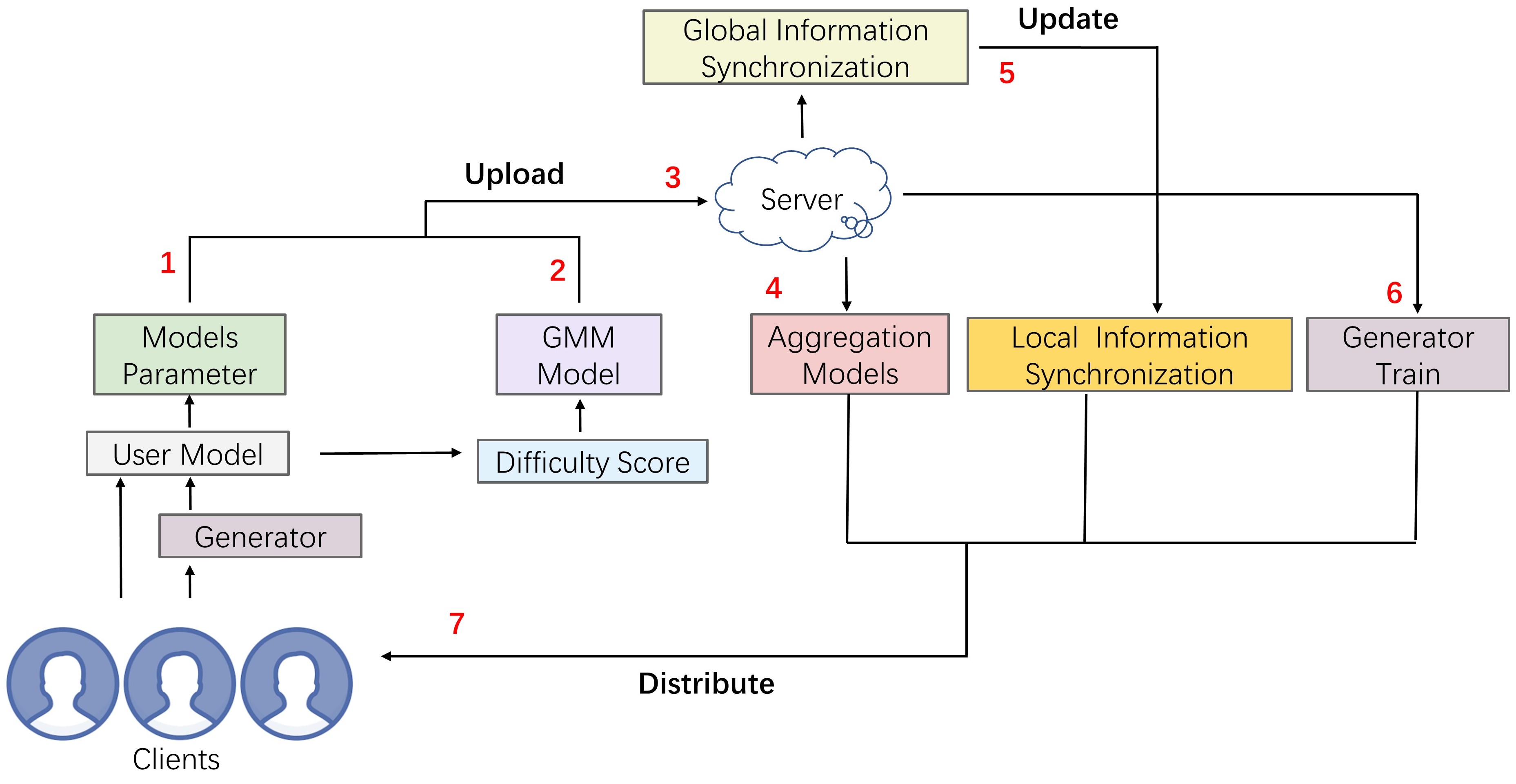}
\caption{Illustration of Federated Multi-phase Curriculum Learning.
The first step involves training the original data and the data generated by the generator for the client. The second step encapsulates the difficulty score during training into the GMM. The third step uploads the results of the first and second steps to the server. The fourth step consists of server aggregation of the master model. The fifth step updates the local information synchronization strategy through the global information synchronization strategy. The sixth step trains the generator model. The seventh step has the server distributing the client main model, generator model, and local synchronization strategy.}
\label{Mod}
\end{figure}
To address the issue of model drift in asynchronous federated learning (FL) approaches, we propose a multi-phase curriculum learning approach called FedCL that deals with user heterogeneity synchronously. FedCL comprises two components: clients and the server.

For each client, the training process consists of four parts. First, each client employs the curriculum learning method to train its model and collect the local difficulty score, which determines the progress of local model convergence. Second, to ensure privacy protection, the local difficulty score is encapsulated using GMM. Third, to facilitate global data distribution synchronization and reduce model drift during local training, we introduce a generator model based on global data and difficulty distribution. The generator model is trained only on the server and does not impose additional training costs on clients. Finally, the client uploads its label counter, task model parameters, and GMM parameters to the server.

For the server, it utilizes the approximate local difficulty score distribution of GMMs uploaded by clients to obtain the global difficulty score distribution, which updates the global information synchronization strategy. Then, each client's local information synchronization strategy is updated based on the global information synchronization strategy, determining if further training is needed for each client. Next, the server performs standard model aggregation and trains the generator model. Finally, the aggregated models and local information synchronization strategy are distributed to the selected clients requiring further training.

The overview of the training procedure is shown in Figure \ref{Mod} and Figure \ref{IID}. We will explain our approach in detail from three perspectives: Global Difficulty Distribution, Generator Learning, and Synchronization Strategy.

\subsection{Global Difficulty Distribution}
\label{DD}
In order to protect the privacy of local difficulty scores during the uploading process from clients to the server, each client $k$ will upload its GMM parameters $(\boldsymbol{\mu}_k,\boldsymbol{\sigma}^2_k,\boldsymbol{\alpha}_k)$ to the server.
For the server, after receiving the GMM parameters from all clients, it wants to generate a global distribution $\beta$ that can fit the difficulty scores of all clients. However, this distribution $\beta$ is hard to obtain. To solve it, the server collects a certain number of samples ($N_k$ samples) for each client $k$ according to the GMM determined by its uploaded parameters $(\boldsymbol{\mu}_k,\boldsymbol{\sigma}^2_k,\boldsymbol{\alpha}_k)$. Thus, we can get a collection of difficulty samples generated from GMMs of all clients. Its total number of $N=\sum_{k=1}^KN_k$. Then, we sort these $N$ difficulty samples from low to high to form a sorted list $\mathcal{N}$. When we want to generate a difficulty sample according to this global distribution $\beta$, we first uniformly generate an integer number $r$ from $1$ to $N$, then return the $r$-th element in the sorted list $\mathcal{N}$. This process can be used to approximately generate difficulty samples from $\beta$.

\subsection{Generator learning}

The process of Knowledge Distillation (KD) involves a student model learning from one or more teacher models. By leveraging KD, the student model can be compressed while simultaneously inheriting knowledge that has been distilled from the teacher(s). One popular approach to Data-Free KD is through the implementation of Generative Adversarial Networks (GANs). Specifically, a generator is trained to produce imitated training data, which then enhances the original training dataset used during the Knowledge Distillation process. The KD has been used in FL to address user heterogeneity \cite{li2020federated} \cite{li2020federated2} by training each local model $k$, parameterized by $\boldsymbol{\omega}_k=(\boldsymbol{\omega}^{f}_{k},\boldsymbol{\omega}^{h}_{k})$ as the teacher, then it is combined into the global (student) model, parameterized by $\boldsymbol{\omega}$.
To improve the generalization performance, it leverages a proxy dataset $\hat{\mathcal{D}}_p$ to minimize the discrepancy, measured by using Kullback-Leibler (KL) divergence shown as follows:
\begin{align}
    \min\limits_{\boldsymbol{{\omega}}}\mathbb{E}_{x \sim  \hat{\mathcal{D}}_p}&{\rm KL}\left[\sigma\left(\frac{1}{K}\sum\nolimits_{k=1}^{K}  g(f({x}; \boldsymbol{\omega}^{f}_{k});\boldsymbol{{\omega}}{^{h}_{k}})\right)\right.\nonumber\\
    &\left.||\sigma(g(f({x} ; \boldsymbol{\omega}^{f});\boldsymbol{{\omega}}{^{h}}) )\right],
\end{align}
where $\sigma(\cdot)$ is an non-linear activation and $g(\cdot)$ is the logits output of an predictor $h$, namely $h(z;\boldsymbol{\omega}^{h})=\sigma(g(z;\boldsymbol{\omega}^{h}))$. However, this requires a proxy dataset $\hat{\mathcal{D}}_p$, whose quality plays an important role in the final performance. Subsequently, we will demonstrate how KD can be made feasible for Synchronous FL in a data-free manner.

The first part of generator learning is to extract knowledge. To generate samples that align with the overall data distribution and satisfy difficulty distribution requirements, it's necessary to extract knowledge about both the global data distribution $\mathcal{D}$ and the difficulty distribution $\beta$.  we propose learning the features of the conditional distribution $Q^*:\mathcal{Y} \rightarrow \mathcal{X}$, which is consistent the ground-truth data distribution and current difficulty distribution:
 \begin{equation}\label{eq13}
Q^*=\mathop{\arg\max}_{{Q: \mathcal{Y} \rightarrow \mathcal{X}}}\mathbb{E}_{y \sim p(y)} \mathbb{E}_{x \sim Q({x}|y,\beta)}[\log p(y|x)],
\end{equation}
where $p(y)$ is the ground-truth prior and $p(y|x,\beta)$ is the posterior distribution of the target labels given a difficulty distribution $\beta$. To make Eqn. (\ref{eq13}) solvable, we replace $p(y)$ and $p(y|x)$ with their empirical approximations. To approximate $p(y)$, we have
\begin{equation}
\hat{p}(y) \propto \sum_k \mathbb{E}_{x\sim \hat{\mathcal{D}}_k}\left[\mathrm{I}\left(c^*(x)=y\right)\right] ,
\end{equation}
where $\mathrm{I}(\cdot)$ is an indicator function and $\hat{p}(y)$ can be obtained by requiring the training label counts from clients during the model uploading phase. Meanwhile, to model $p(y|x)$, we can use the ensemble wisdom from user models as follow:
\begin{equation}
\log \hat{p}(y \mid x) \propto \frac{1}{K} \sum_{k=1}^K \log p\left(y|x; \boldsymbol{\omega}_{k}\right).
\end{equation}

Directly optimizing the Eqn. (\ref{eq13}) by using the above approximations, it is prone to overload when $\mathcal{X}$ is high-dimensional and may also reveal information about the user's data profile. A feasible method is to exploit an induced distribution $G^*:\mathcal{Y}\rightarrow\mathcal{Z}$ over the latent space, which is more compact than the original space and provides the privacy protection. That is
\begin{equation}
    G^*=\mathop{\arg\max}_{{G: \mathcal{Y} \rightarrow \mathcal{Z}}}\mathbb{E}_{y \sim \hat{p}(y)}\mathbb{E}_{z \sim G({z}|y,\beta)}\sum_{k=1}^K \log p\left(y|z; \boldsymbol{\omega}_{k}^h\right).
\end{equation}
Following the above explanation, we can learn a conditional generator $G$ parameterized by $\boldsymbol{\theta}$ to optimize $J(\boldsymbol{\theta})$, where we have $\min _{\boldsymbol{\theta}} J(\boldsymbol{\theta})=$
\begin{equation}
\mathbb{E}_{y \sim \hat{p}(y)} \mathbb{E}_{z \sim G_{\boldsymbol{\theta}}(z| y,\beta)}
\left[l^{cl}\left(\sigma\left(\frac{1}{K} \sum_{k=1}^{K} g\left(z ; \boldsymbol{\omega}^{h}_k\right)\right), y\right)\right].
\label{opt}
\end{equation}
The conditional sample $z$ is based on the condition of $y$ and $\beta$, and optimizing Eqn. (\ref{opt}) only requires access to the predictor modules $\boldsymbol{\omega}^{h}_k$ of user models. Specifically, to enable diversified outputs from $G(\cdot|y,\beta)$, we introduce a noise vector $\epsilon \sim \mathcal{N}(0, I)$ to the generator, which is resemblant to the re-parameterization technique proposed by prior art \cite{kingma2013auto}, so that $z \sim G_{\boldsymbol{\theta}}(\cdot|y,\beta) \equiv G_{\boldsymbol{\theta}}(y,\beta, \epsilon|\epsilon \sim\mathcal{N}(0, I))$. 

Given arbitrary target labels $y$, the proposed generator can yield feature representations $z \sim G_{\boldsymbol{\theta}}(\cdot|y,\beta)$ that induce ideal predictions from the ensemble of user models. In other words, the Generator for Global data and difficulty Distribution can be seen as an induced image with a consensual distribution that is consistent with the user data and difficulty distribution from a global perspective. It generates samples that reflect both the global data distribution and the difficulty level of each user's data, thereby achieving the goal of personalized learning in federated learning.

The second part of generator learning is to distill knowledge. The learned generator $G_{\boldsymbol{\theta}}$ is then sent to local clients so that each local model can sample from $G_{\boldsymbol{\theta}}$ to obtain an enhanced representation $z\sim G_{\boldsymbol{\theta}}(\cdot|y,\beta)$ over the feature space. Thus, the objective of the local model parameterized by $\boldsymbol{\omega}_{k}$ is changed to maximize the probability that it will produce an ideal prediction for the enhanced samples. The objective of each local client $k$ can be defined as $\min _{\boldsymbol{\omega}_{k}} J(\boldsymbol{\omega}_{k})=$
\begin{equation}
\hat{\mathcal{L}}^{cl}_{k}\left(\boldsymbol{\omega}_k\right)+\hat{\mathbb{E}}_{y \sim \hat{p}(y), z \sim G_{\boldsymbol{\theta}}(z|y,\beta)}
\left[l^{cl}\left(h\left(z ; \boldsymbol{\omega}_{k}^{h}\right),y\right)\right],
\label{dis}
\end{equation}
where we have
\begin{equation}
\hat{\mathcal{L}}_{k}^{cl}(\boldsymbol{\omega}_k)=
\frac{1}{|\hat{\mathcal{D}}_k|} \sum_{x_i \in \hat{\mathcal{D}}_k} l^{cl}(h(f(x_{i} ; \boldsymbol{\omega}^{f}_k);\boldsymbol{{\omega}}{^{h}_k}), c^{*}(x_{i})).
\end{equation}
It is the empirical risk given a local dataset $\hat{\mathcal{D}}_k$. To this end, we propose an approach to realize knowledge transfer without data by interactively learning a generator that mainly relies on global data and difficulty distribution. The generator can then be used to transmit global knowledge to local users. 

In summary, the Generator for global data and difficulty distribution is a useful approach for inducing a consensusal distribution that aligns with the global data distribution and the difficulty level of each user's data. It essentially generates synthetic samples that reflect both of these factors, thus enabling personalized learning in the context of federated learning. This is achieved by leveraging the difficulty distribution to control the sampling process and ensure that the generated samples are representative of each user's data distribution. As a result, the generated samples can be used to augment the training data of individual users, thereby improving the accuracy and generalization of the federated model. Additionally, the generator can be trained using a variety of techniques, including GANs, VAEs, or other generative models, depending on the specific requirements of the application. Overall, the Generator for global data and difficulty distribution is a powerful tool for addressing the challenges of federated learning and advancing the state-of-the-art in this area.

\subsection{Synchronization Strategy}

We propose a synchronization strategy that capitalizes on the global difficulty score to transform asynchronous federated learning (FL) into synchronous FL. The global information synchronization strategy seeks to harmonize the global learning state, ensuring that all clients learn from a congruous global perspective and update their local models in a coordinated manner, ultimately augmenting the overall performance of the FL algorithm.

Global information synchronization encompasses acquiring the global difficulty score and distributing local information synchronization to each user. For the global difficulty score, the server samples a number of difficulty samples based on the Gaussian Mixture Models (GMMs) returned by clients to obtain an approximation of the global difficulty score distribution $\beta$, as delineated in Section \ref{DD}. For local information synchronization, we define a temperature $T \in (0, 1)$ as the threshold, signifying the state of global synchronization. We then define a decreasing sorted list $\mathcal{S}={T_{1}, T_{2},\cdots,T_{Z}}$ as all training states, comprising a sequence of thresholds such that $T_{1}\geq T_{2}\geq\cdots\geq T_{Z}$. As expounded in Section \ref{DD}, the server can acquire a sorted list $\mathcal{N}$ by sampling according to all local GMMs, mirroring the global difficulty distribution $\beta$. Given a threshold $T_z\in\mathcal{S}$, the server returns the $T_z\cdot N$-th element (difficulty sample) in $\mathcal{N}$, denoted as $\beta_{T_z}$ for convenience. Predicated on the $(\boldsymbol{\mu}_k,\boldsymbol{\sigma}^2_k,\boldsymbol{\alpha}_k)$ returned by client $k$, the server reassembles a set of difficulty score samples $\mathcal{N}_k$ with $|\mathcal{N}k|=N_k$. We then compare the samples in $\mathcal{N}k$ with the global $\beta{T_z}$. If the number of samples in ${x\in\mathcal{N}k|x\leq\beta{T_z}}$ is greater than ${v}\cdot N_k$, where $v$ is a temperature coefficient, it suggests that the majority of samples in client $k$ satisfy the current global state prerequisites. The server then issues a freeze instruction to the corresponding client, delineating a set $M_f$ consisting of all frozen clients and a set $M_l$ encompassing all unfrozen clients. Clients in $M_l$ necessitate further training. When all local models are frozen, it denotes the completion of learning under the current threshold $T_z$, progressing to the next threshold $T{z+1}$, and unfreezing all clients at that point.

\begin{algorithm}[!t]
\caption{FedCL}  
\label{FedCLAlg}  
\begin{algorithmic}[1]  
\REQUIRE
Tasks $\{{T}_k\}_{k=1}^K$ and Difficulty score $\bar{l}^{c l}$\;

Global difficulty distribution${\mathcal{D}}^{cl}$ and local difficulty distribution ${\mathcal{D}}^{cl}_k\sim(\boldsymbol{\mu}_k,\boldsymbol{\sigma}^2_k,\boldsymbol{\alpha}_k)$;

Global parameter $\boldsymbol{\omega}$ and Local parameters $\{\boldsymbol{\omega_k}\}_{k=1}^K$\;

Generator parameter $\boldsymbol{\theta}$\; 

$\hat{p}(y)$ and global difficulty score distribution $\beta$ are uniformly initialized\;

Global threshold $\mathcal{S}=\{{T}_{z}\}_{z=1}^Z$;

Local maximum steps $T$ and batch size $B$\;

Learning rate $\gamma$ and $\alpha$\;
   
\FOR{each training state ${T}_{z}\in\mathcal{S}$}
\STATE $M_f\leftarrow\emptyset$, $M_l\leftarrow[K]$;
\WHILE{$M_l\neq\emptyset$}
\STATE Send global $\boldsymbol{\omega}$, $\boldsymbol{\theta}$, $\hat{p}(y)$, $\beta$ to clients in $M_l$;
\FOR{each client $k\in M_l$ in parallel}
    \STATE $\boldsymbol{\omega}_k \leftarrow \boldsymbol{\omega}$;
        \FOR{$t=1$ \KwTo $T$}
        \STATE $\left\{x_i, y_i\right\}_{i=1}^B \sim \mathcal{T}_k$;
        \STATE $\left\{\hat{z}_i \sim G_{\boldsymbol{\theta}}(\cdot|\hat{y}_i,\beta), \hat{y}_i \sim \hat{p}(y)\right\}_{i=1}^B$;
        \STATE Update label counter $c_k$;
        \STATE $\boldsymbol{\omega}_{k}\leftarrow {\boldsymbol{\omega}{_k}-\gamma\nabla_{\boldsymbol{\omega}_{k}}{J}\left(\boldsymbol{\omega}{_k}\right)}$ in Eqn. \eqref{dis};
        \ENDFOR
    \STATE $\mathcal{D}_k^{c l}\leftarrow(\boldsymbol{\mu}_k,\boldsymbol{\sigma}^2_k,\boldsymbol{\alpha}_k)$;
    \STATE Sends $\boldsymbol{\omega}_{k}$, $\mathcal{D}_k^{c l}$, and $c_k$ to the server;
\ENDFOR
\STATE Server updates $\boldsymbol{w} \leftarrow \frac{1}{|{K}|} \sum_{k \in [K]} \boldsymbol{w}_k$, and $\hat{p}(y)$ based on $\{c_k\}_{k\in [K]}$;
\STATE $\boldsymbol{\theta} \leftarrow \boldsymbol{\theta} - \alpha \nabla_{\boldsymbol{\theta}}J(\boldsymbol{\theta})$ as in Eqn. \eqref{opt};
\STATE $\beta\leftarrow\{\mathcal{D}_k^{c l}\}_{k=1}^K$; 
\STATE Generate a difficulty sample set $\mathcal{N}$ based on $\beta$ and determine $\beta_{T_z}$;
\FOR{each client $k\in M_l$}
    \STATE Collect a set of difficulty samples $\mathcal{N}_k$ on $\mathcal{D}_k^{c l}$;
    \IF{$|\{x\in\mathcal{N}_k|x\leq\beta_{T_z}\}|>N_k\cdot{v}$} 
        \STATE $M_f\leftarrow M_f\cup\{k\}$;
    \ENDIF
\ENDFOR
\STATE $M_l\leftarrow [K]\backslash M_f$;

\ENDWHILE
\ENDFOR
\end{algorithmic}  
\end{algorithm}

Therefore, the server aggregates an approximate unbiased model, as shown in Figure \ref{Mod}. 
For each interaction between the server and clients, the server can determine the current frozen set $M_f$ and unfrozen set $M_l$. Then, the global parameters should be aggregated as
\begin{equation}
\boldsymbol{\omega}
=\frac{1}{K}\left[
\sum_{k \in M_f} \boldsymbol{\omega}_k+
\sum_{k \in M_l} \boldsymbol{\omega}_k
\right],
\label{fzz}
\end{equation}
where the server applies the parameter averaging technique to update global parameters by averaging the parameters of both $M_f$ and $M_l$. This strategy aims to solve the problem of model drift caused by overfitting personalized user features. Besides, neglecting the aggregation of frozen models can lead to insufficient representation of general features, resulting in poor performance of the model. Therefore, it is essential to aggregate all models, including frozen models, to obtain optimal generalization performance.

\subsection{Algorithm Design}

Our approach can extend any asynchronous FL to synchronous FL, which significantly alleviates the heterogeneity problem under Non-IID data. The pseudocode of our proposed method is shown in Algorithm \ref{FedCLAlg}.

At the beginning of each training state $T_z\in\mathcal{S}$, the $M_f$ and $M_l$ are initialized by the empty set and $[K]$ respectively. For each interaction between the server and clients, the server first sends the aggregated model $\boldsymbol{w}$, parameters of the generator $\boldsymbol{\theta}$, and $\hat{p}(y)$, and global difficulty score distribution $\beta$ to the selected client in $M_l$. Each selected client $k\in M_l$ trains its local model based on its original data, using the global learning state $\beta$ and generated data $G_{\boldsymbol{\theta}}(\cdot|\hat{y}i,\beta)$ for global information synchronization strategy. The difficulty level for each training state $\beta$ is determined by the main server. Then, it sends its updated weight $\boldsymbol{\omega}{k}$, parameters of difficulty score distribution $\mathcal{D}k^{c l}$ and local label counter $c_k$ to the server. This process is executed on the client side, which is shown in lines 5 to 15 of Algorithm \ref{FedCLAlg}. After the server receives the feedback from clients, it averages the parameters of all local models to update the global model $\boldsymbol{w}$. The parameters of the generator $\boldsymbol{\theta}$ are also updated based on the updated global model. Finally, the server estimates the global difficulty score distribution $\beta$ and determines $\beta_{T_z}$. Then, it updates $M_f$ and $M_l$ based on $\beta_{T_z}$. This process is executed on the server side, which is shown in lines 16 to 26 of Algorithm \ref{FedCLAlg}. The process continues until all clients have been frozen, which implies that all local models are under the same learning state, at which point we proceed to the next training state.

\section{EXPERIMENTS}

\begin{table}[!t]
	\caption{The Configurations for Experiments}
	\centering
			\resizebox{1\linewidth}{!}{%
			\begin{tabular}{lllcc}
				\hline
				\multicolumn{2}{l}{Approach} & \multicolumn{2}{|l|}{Hyperparameter} & Value \\ \cmidrule{1-5}
				\multicolumn{2}{l}{\multirow{8}{*}{Shared Parameters}}                                             & \multicolumn{2}{|l|}{Learning Rate}                             & 0.01                 \\ \cmidrule{3-5}
				\multicolumn{2}{l}{}                                                                               & \multicolumn{2}{|l|}{Optimizer}                                 & SGD                  \\ \cmidrule{3-5}
				\multicolumn{2}{l}{}                                                                               & \multicolumn{2}{|l|}{Local Update Steps}                        & 20                   \\ \cmidrule{3-5}
				\multicolumn{2}{l}{}                                                                               & \multicolumn{2}{|l|}{Batch Size}                                & 32                   \\ \cmidrule{3-5}
				\multicolumn{2}{l}{}                                                                               & \multicolumn{2}{|l|}{CELEBA Communication rounds}               & 100                  \\ \cmidrule{3-5}
				\multicolumn{2}{l}{}                                                                               & \multicolumn{2}{|l|}{Other's Communication rounds}              & 200                  \\ \cmidrule{3-5}
				\multicolumn{2}{l}{}                                                                               & \multicolumn{2}{|l|}{\# of total users}                         & 20                   \\ \cmidrule{3-5}
				\multicolumn{2}{l}{}                                                                               & \multicolumn{2}{|l|}{\# of active users}                        & 10                   \\ \cmidrule{1-5}
				\multicolumn{2}{l}{\multirow{4}{*}{\begin{tabular}[l]{@{}l@{}}FEDGEN \& FEDCL\\ Shared \\ Parameterss\end{tabular}}} & \multicolumn{2}{|l|}{Generator Optimizer}                       & adam                 \\ \cmidrule{3-5}
				\multicolumn{2}{l}{}                                                                               & \multicolumn{2}{|l|}{Generator learning rate}                   & 10$^{-4}$                   \\ \cmidrule{3-5}
				\multicolumn{2}{l}{}                                                                               & \multicolumn{2}{|l|}{Generator inference size}                  & 128                  \\ \cmidrule{3-5}
				\multicolumn{2}{l}{}                                                                               & \multicolumn{2}{|l|}{User distillation batch size}              & 32                   \\ \cmidrule{1-5}
				
				\multicolumn{2}{l}{\multirow{2}{*}{FEDCL CL Settings}}                                             & \multicolumn{2}{|l|}{Curriculum Completed Threshold}                             & 80\%                \\ \cmidrule{3-5}
				\multicolumn{2}{l}{}                                                                               & \multicolumn{2}{|l|}{Learning levels of CL}                                 & [0.3,0.6,0.9]                  
				\\ \cmidrule{1-5}
				
				\multicolumn{2}{l}{\multirow{2}{*}{FEDDISTILL \& FEDDISTILL$^+$}}                                     & \multicolumn{2}{|l|}{\multirow{2}{*}{Distillation Coefficient}} & \multirow{2}{*}{0.1} \\
				\multicolumn{2}{l}{}                                                                               & \multicolumn{2}{|l|}{}                                          &                      \\ \cmidrule{1-5}
				\multicolumn{2}{l}{FEDPROX}                                                                        & \multicolumn{2}{|l|}{Proximal Coefficient}                      & 0.1                  \\ \cmidrule{1-5}
		\end{tabular}
	}
	\label{Experiment Configurations}
\end{table}

\begin{table*}[!h]
	\caption{Performance Overview under Different Data Heterogeneity Settings.}
	\centering
        \scriptsize
        \setlength{\tabcolsep}{5.2mm}{
        \renewcommand\arraystretch{1.2}
			\begin{tabular}{l|c|c|c|c|c|c|c}
				\hline
				\multicolumn{8}{c}{\textbf{Top-1 Test Accuracy}} \\ \hline
				Dataset  & Setting & FEDAVG  & FEDPROX  & FEDENSEMBLE & FEDDISTILL & FEDGEN & FEDCL   \\  \cmidrule{1-8}
				\multirow{3}{*}{CELEBA}                                                 
                    & $r=5/10$  & 86.85±0.43 & 87.34±0.43 & 87.94±0.31 & 76.73±1.21 & 89.45±0.38 & \textbf{90.65±0.42} \\ \cmidrule{2-8}
				& $r=5/25$  & 89.19±0.24 & 88.69±0.22 & 90.13±0.28 & 75.11±1.62 & 89.26±0.42 & \textbf{90.83±0.61} \\ \cmidrule{2-8}
				& $r=10/2$5 & 89.07±0.18 & 88.98±0.37 & 90.04±0.21 & 75.91±1.19 & 90.27±0.44 & \textbf{91.58±0.64} \\ \cmidrule{1-8}
				\multirow{4}{*}{MNIST}                                                  
                    & $\alpha=0.05$  & 86.92±1.56 & 86.73±1.97 & 89.05±1.04 & 71.84±1.46 & 90.04±0.68 & \textbf{91.47±0.81} \\ \cmidrule{2-8}
				& $\alpha=0.1$   & 91.13±0.84 & 90.69±0.99 & 91.60±0.56 & 58.03±0.78 & 95.86±0.05 & \textbf{96.03±0.12} \\ \cmidrule{2-8}
				& $\alpha=1$ & 94.73±0.49 & 94.40±0.37 & 94.77±0.49 & 86.92±1.56 & 97.36±0.14 & \textbf{97.56±0.02} \\ \cmidrule{2-8}
				& $\alpha=10$    & 95.03±0.51 & 94.81±0.50 & 94.98±0.52 & 86.73±1.97 & 97.52±0.04 & \textbf{97.62±0.03} \\ \cmidrule{1-8}
				\multirow{4}{*}{\begin{tabular}[c]{@{}c@{}}EMNIST \\ $T=20$\end{tabular}} 
                    & $\alpha=0.05$ & 63.00±1.06 & 62.75±0.44 & 64.52±0.87 & 47.13±0.14 & 71.98±1.01 & \textbf{73.65±1.23} \\ \cmidrule{2-8}
                    & $\alpha=0.1$ & 73.65±1.23 & 68.55±0.87 & 69.72±0.72 & 45.84±0.23 & 76.05±0.22 & \textbf{78.06±1.25} \\ \cmidrule{2-8}
                    & $\alpha=1$ & 78.08±0.41 & 77.28±0.54 & 78.34±0.41 & 56.12±0.36 & 82.81±0.05 & \textbf{84.21±0.15} \\ \cmidrule{2-8}
                    & $\alpha=10$ & 77.81±0.11 & 77.03±0.04 & 77.83±0.03 & 46.34±0.11 & 82.71±0.09 & \textbf{83.88±0.29} \\ \cmidrule{1-8}
				\multirow{4}{*}{\begin{tabular}[c]{@{}c@{}}EMNIST \\ $T=40$\end{tabular}} 
                    & $\alpha=0.05$ & 68.72±0.60 & 66.85±0.57 & 69.57±0.63 & 49.32±0.13 & 72.29±1.27 & \textbf{74.91±2.55} \\ \cmidrule{2-8}
                    & $\alpha=0.1$ & 75.18±0.54 & 74.05±0.82 & 75.92±0.67 & 48.23±0.42 & 77.43±0.54 & \textbf{79.25±1.53} \\ \cmidrule{2-8}
                    & $\alpha=1$ & 83.79±0.30 & 83.22±0.22 & 82.91±0.15 & 56.83±0.53 & 84.01±0.14 & \textbf{84.77±0.18} \\ \cmidrule{2-8}
                    & $\alpha=10$ & 83.47±0.20 & 82.91±0.15 & 83.62±0.15 & 48.67±0.53 & 84.24±0.27 & \textbf{84.77±0.18} \\ \cmidrule{1-8}
	\end{tabular}
	}
	\label{Experiment Results}
 
\end{table*}

\begin{figure*}[!h]
\centering
	\subfloat[$\alpha=0.05$]{
		\includegraphics[width=0.21\textwidth]{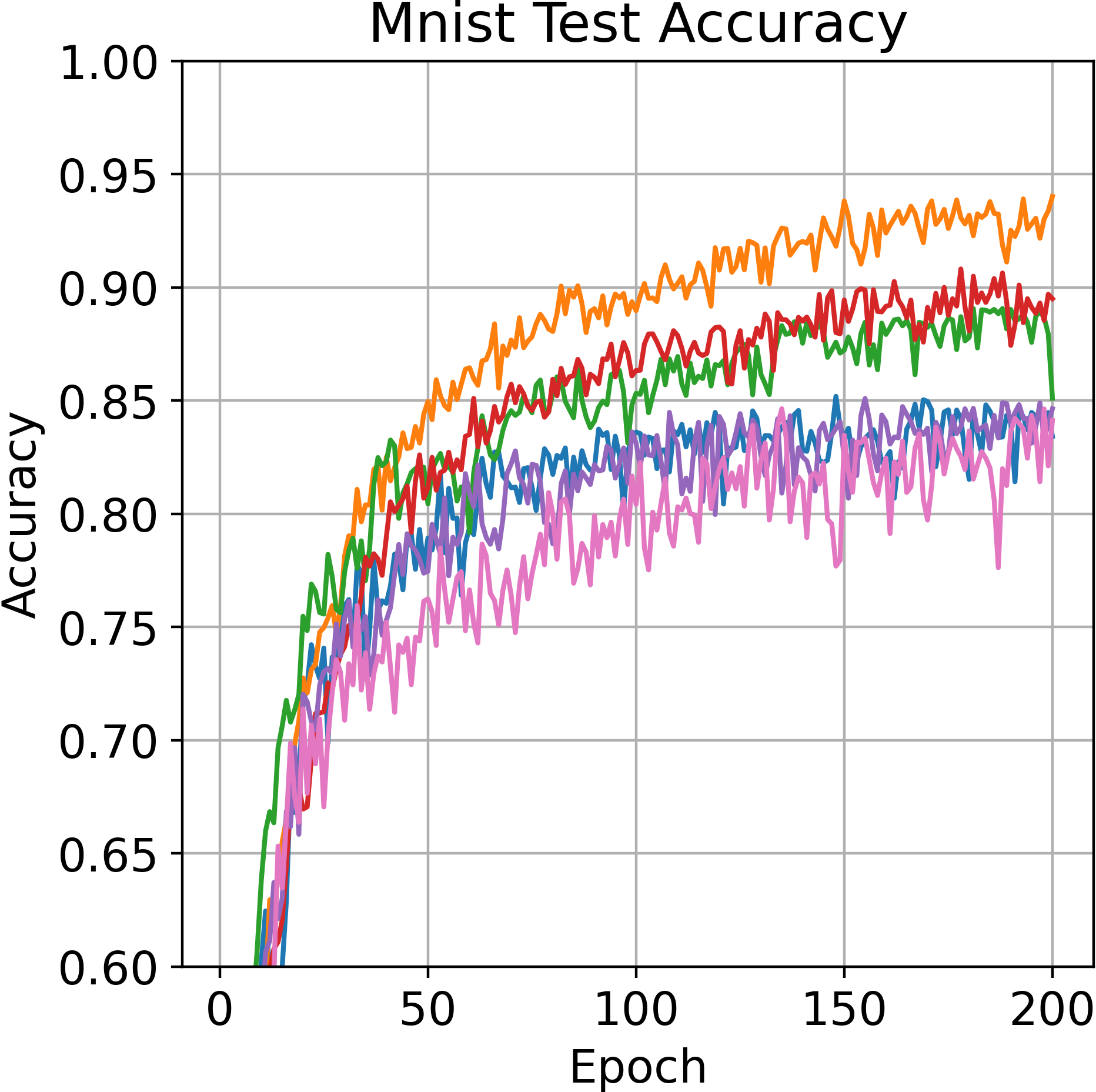}
	}
	\subfloat[$\alpha=0.1$]{
		\includegraphics[width=0.21\textwidth]{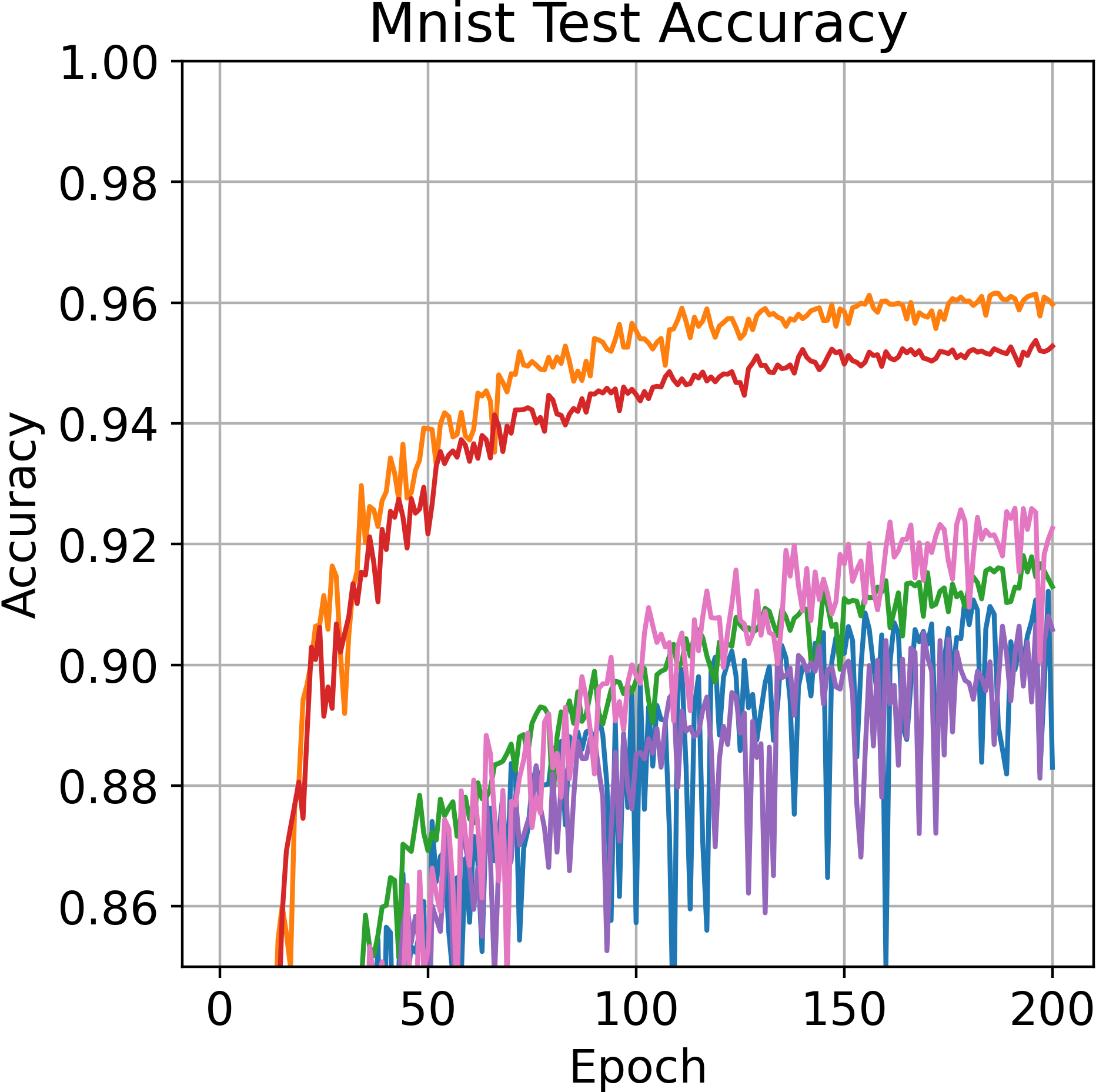}
	}
	\subfloat[$\alpha=1$]{
	    \includegraphics[width=0.21\textwidth]{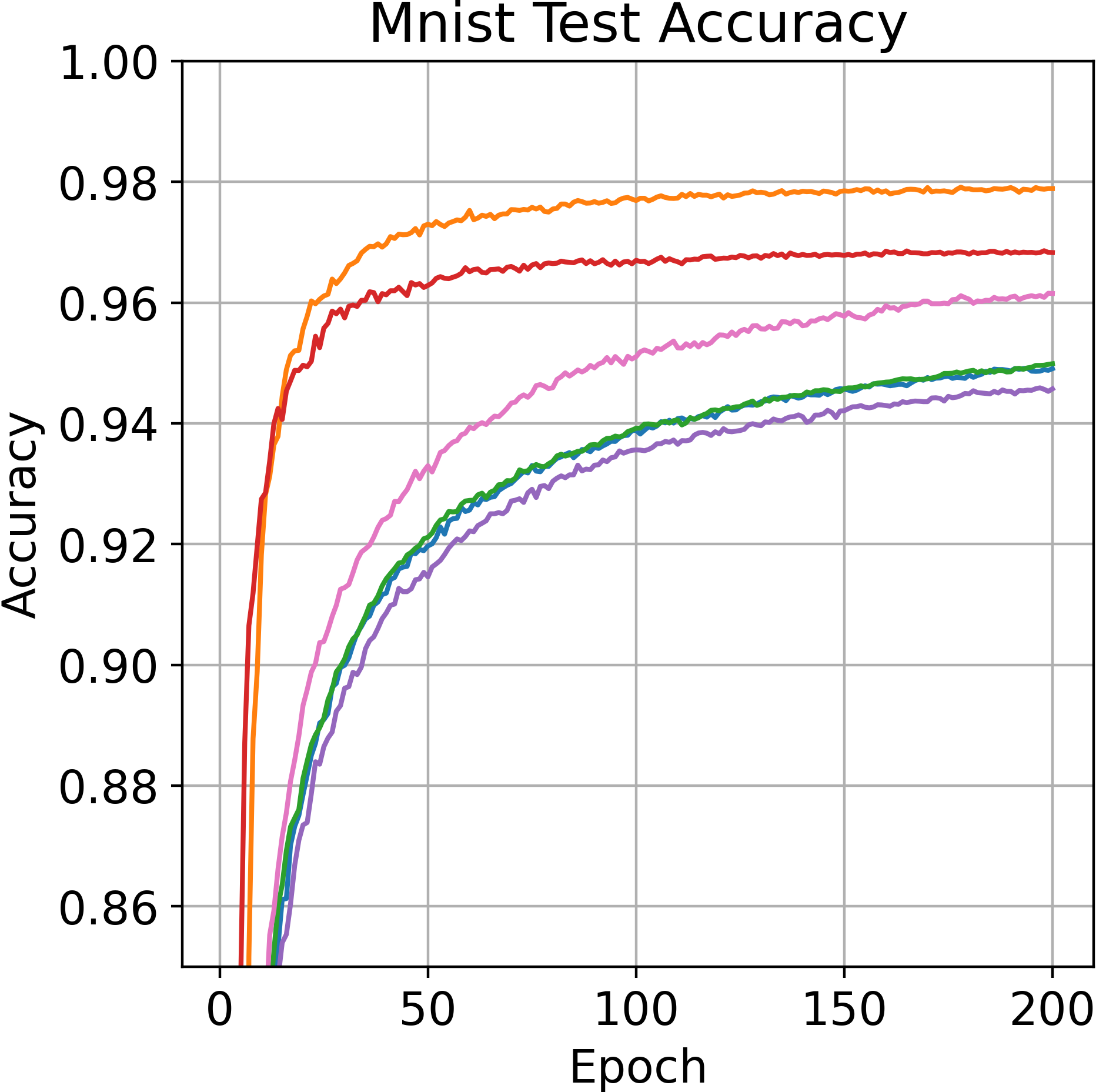}
	}
	\subfloat[$\alpha=10$]{
			\includegraphics[width=0.21\textwidth]{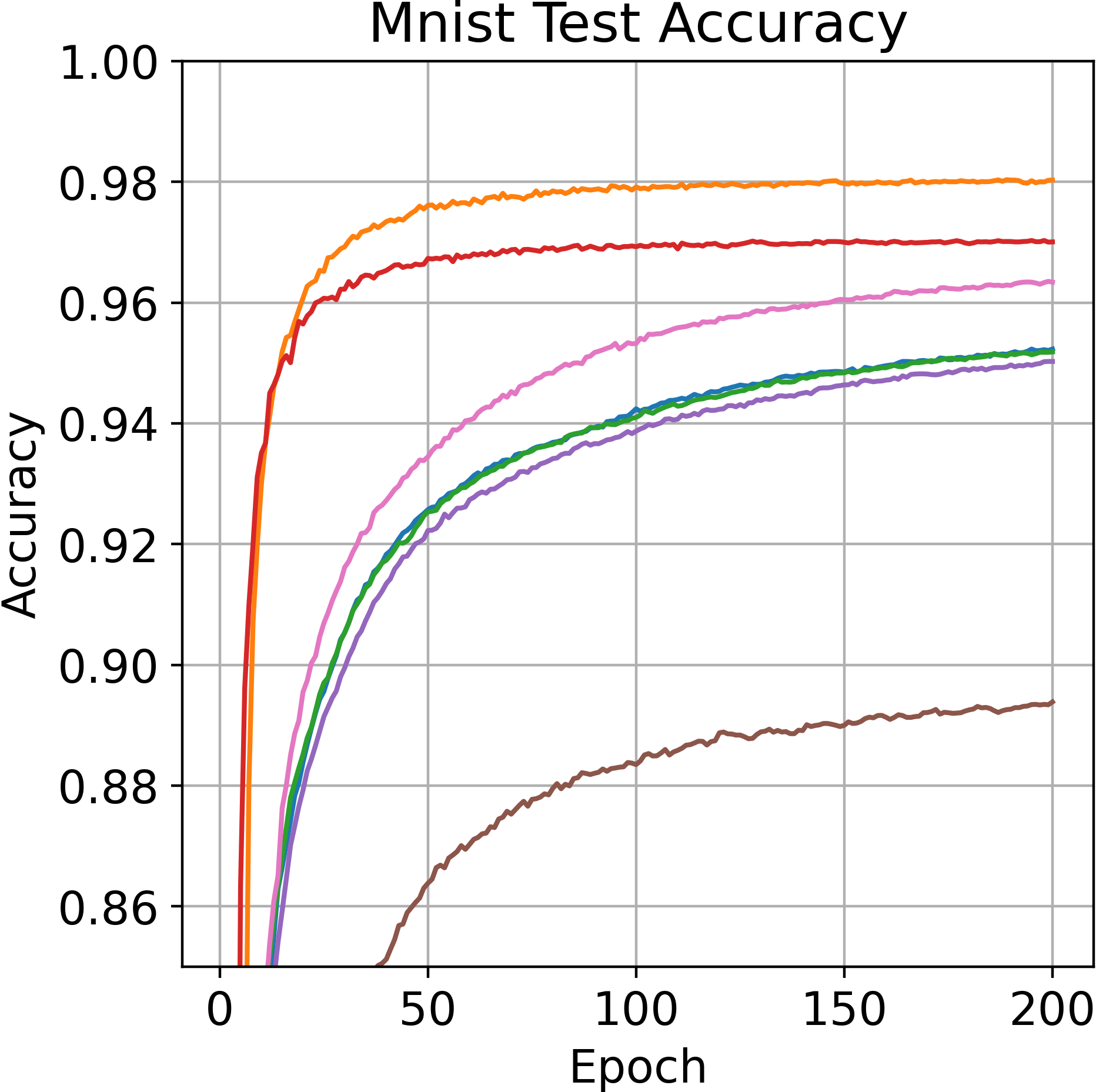}
	}
	\subfloat{
			\includegraphics[width=0.08\textwidth]{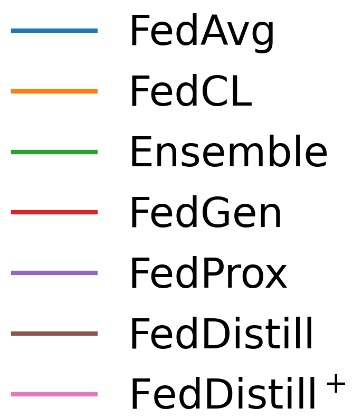}
	}

    \caption{Performance curves on MNIST dataset where a smaller $\alpha$ means larger data heterogeneity.}
    \label{MNIST Result}
\end{figure*}

\begin{figure*}[!h]
\raggedright
	\subfloat[$r=5/10$]
	{
			\includegraphics[width=00.21\textwidth]{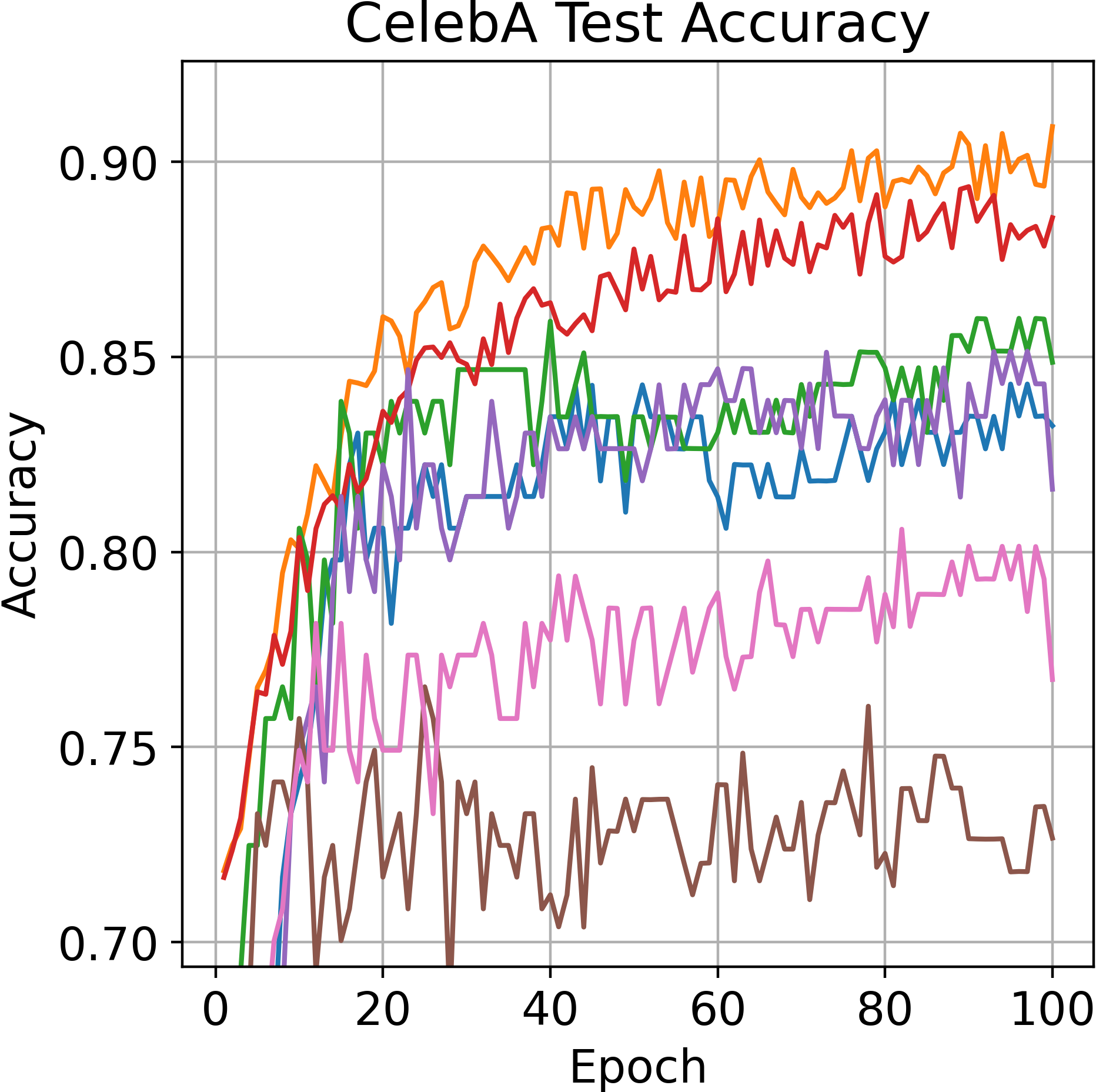}
	}
	\subfloat[$r=5/25$]
	{
		\includegraphics[width=0.21\textwidth]{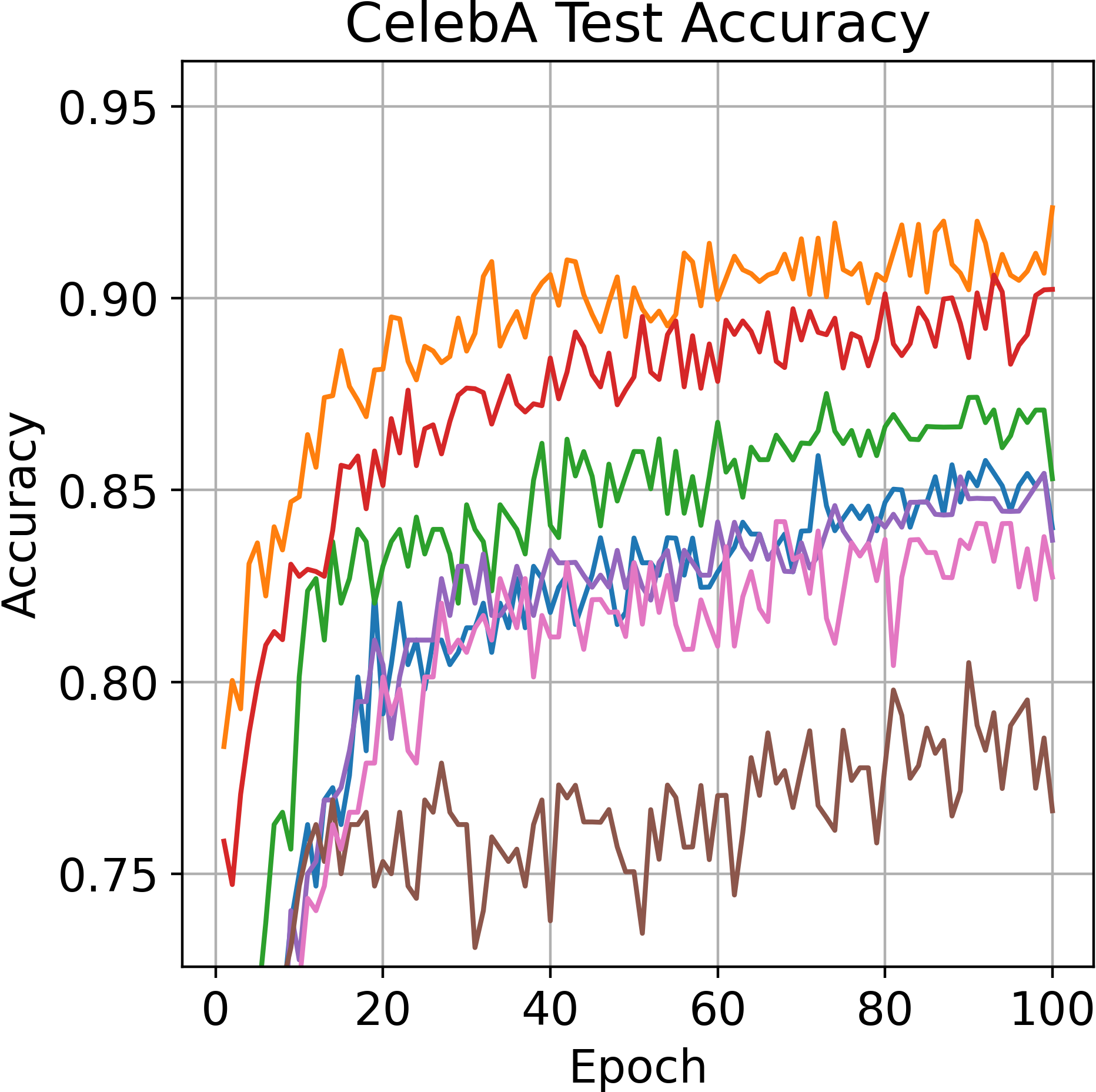}
	}
	\subfloat[$r=10/25$]
	{
			\includegraphics[width=0.21\textwidth]{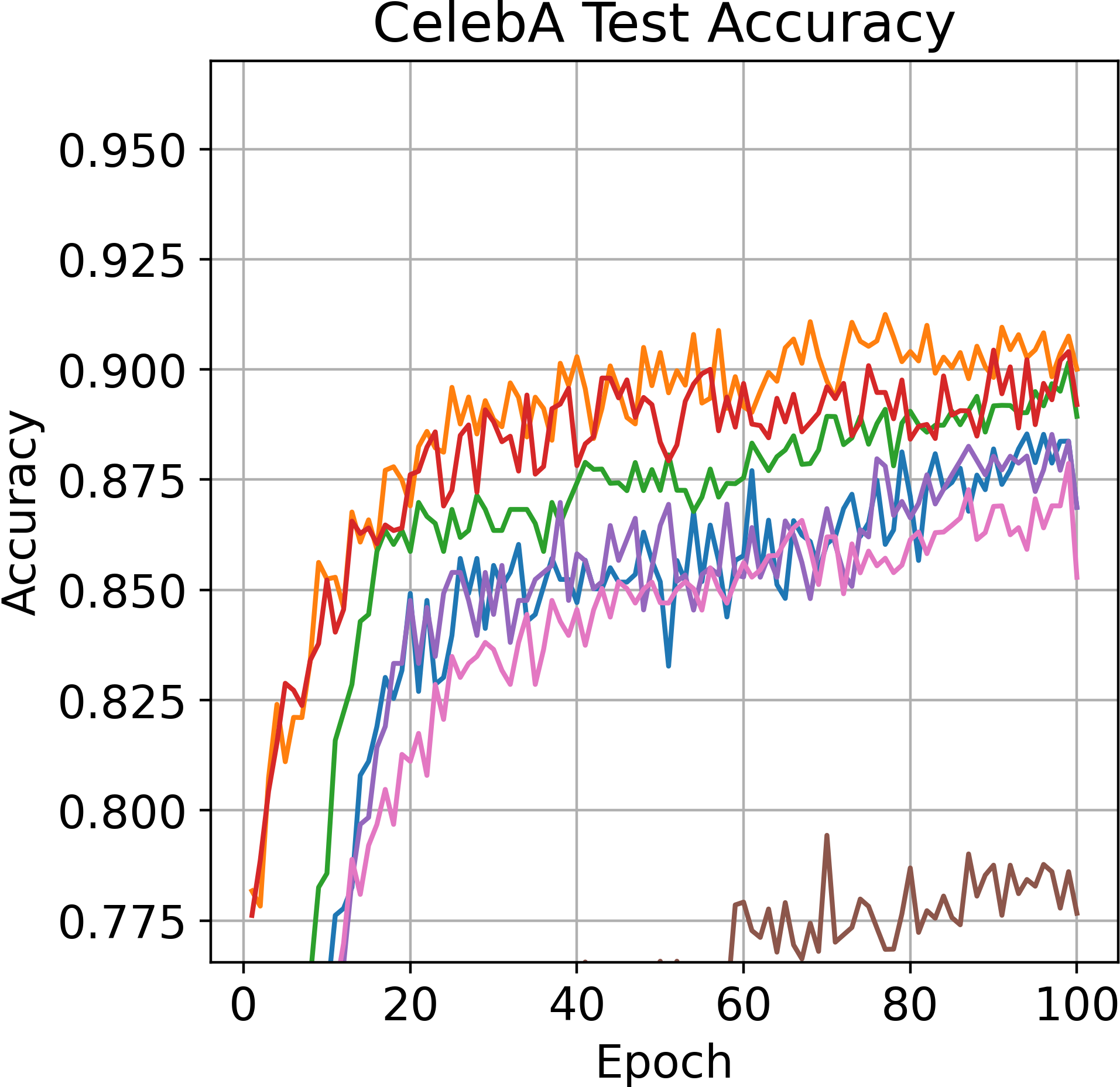}
	}
	\subfloat
    {
		\includegraphics[width=0.08\textwidth]{color-model.png}
    }
	\caption {Performance Curves on CELEBA Dataset}
	\label{CELEBA Result}
\end{figure*}

\begin{figure*}[!h]
    \centering
	\subfloat[$\alpha=0.05, T=20$]
	{
			\includegraphics[width=0.21\textwidth]{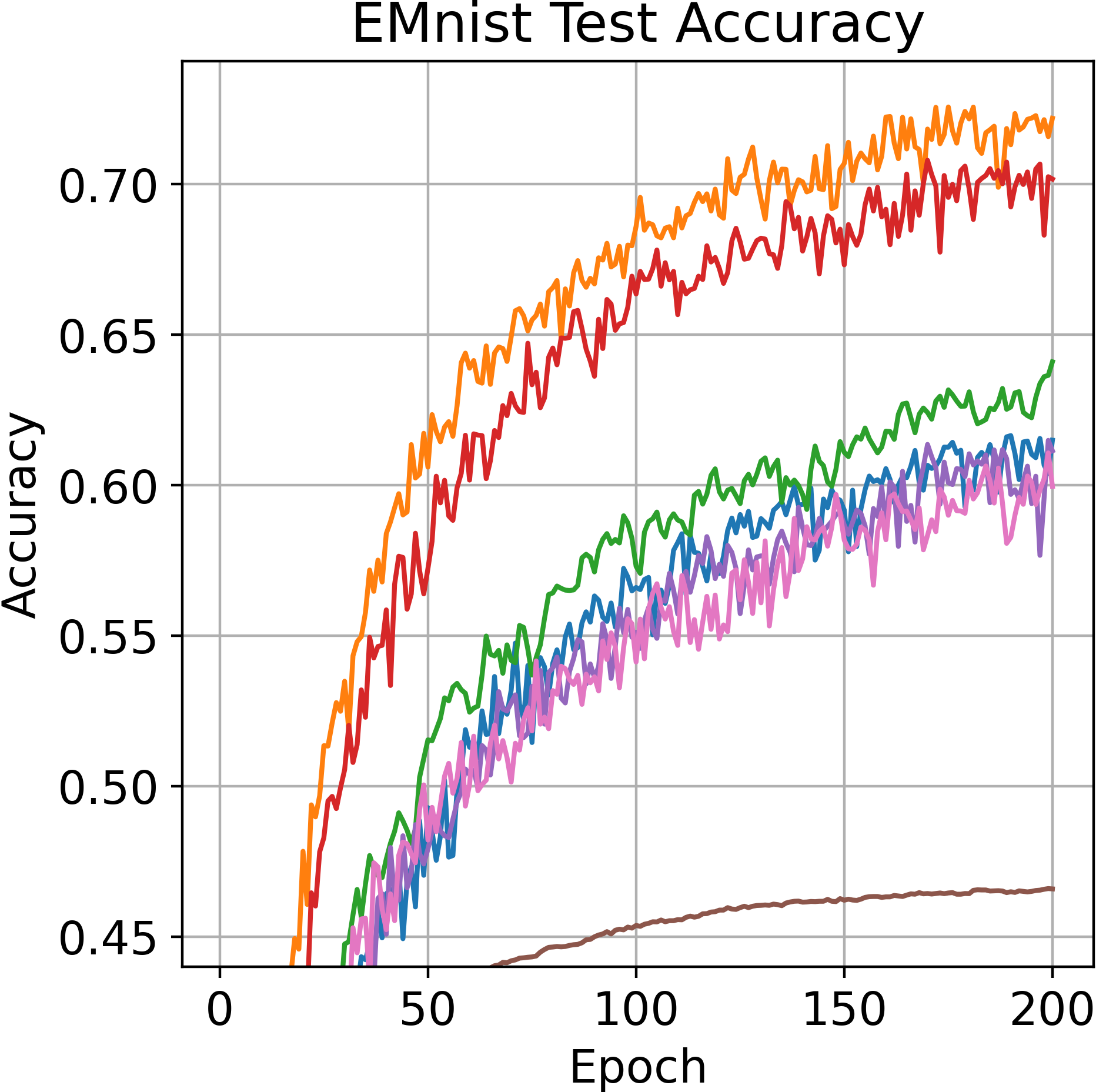}
	}
	\subfloat[$\alpha=0.1, T=20$]
	{
			\includegraphics[width=0.21\textwidth]{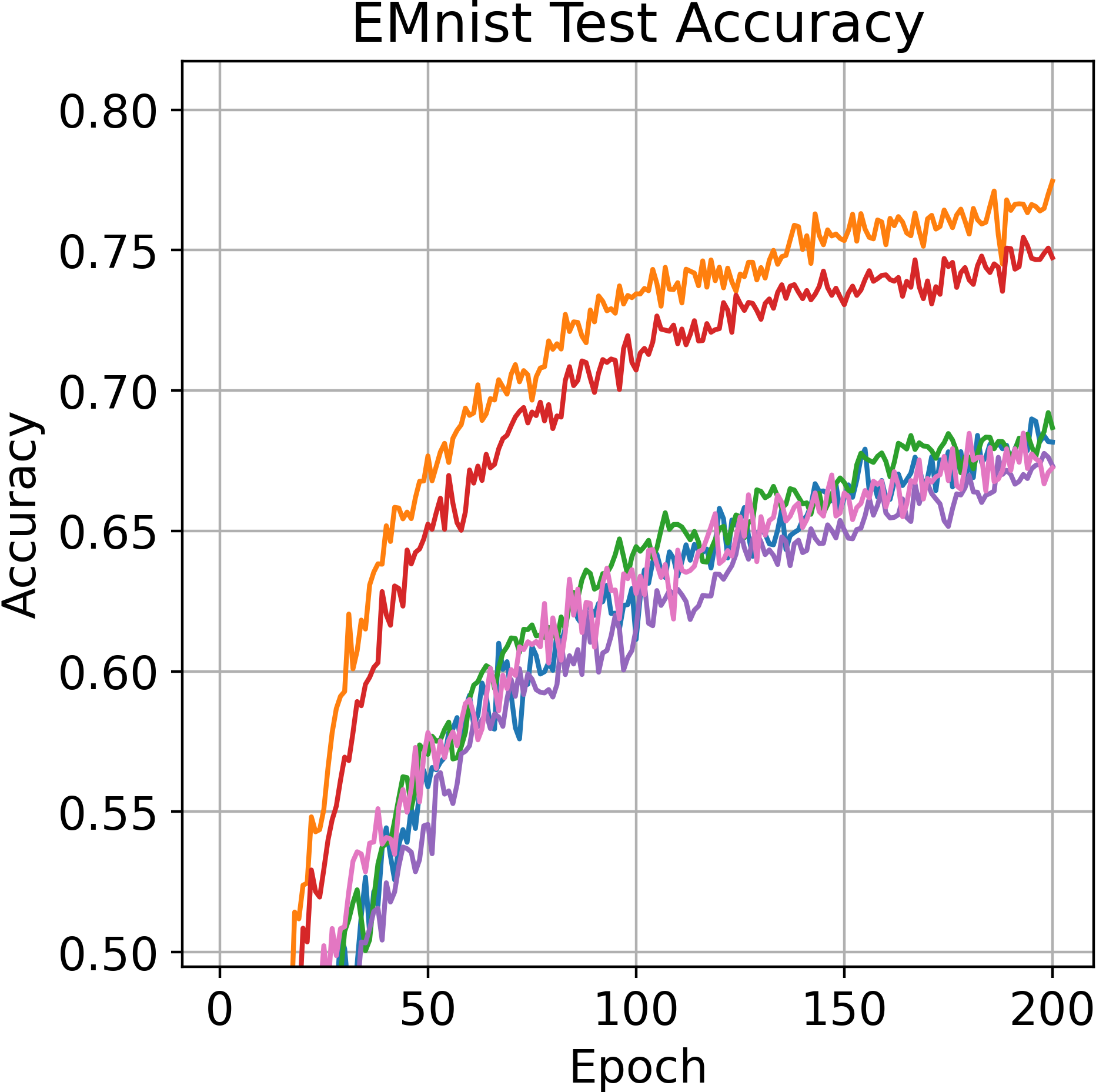}
	}
	\subfloat[$\alpha=1, T=20$]
	{
		\includegraphics[width=0.21\textwidth]{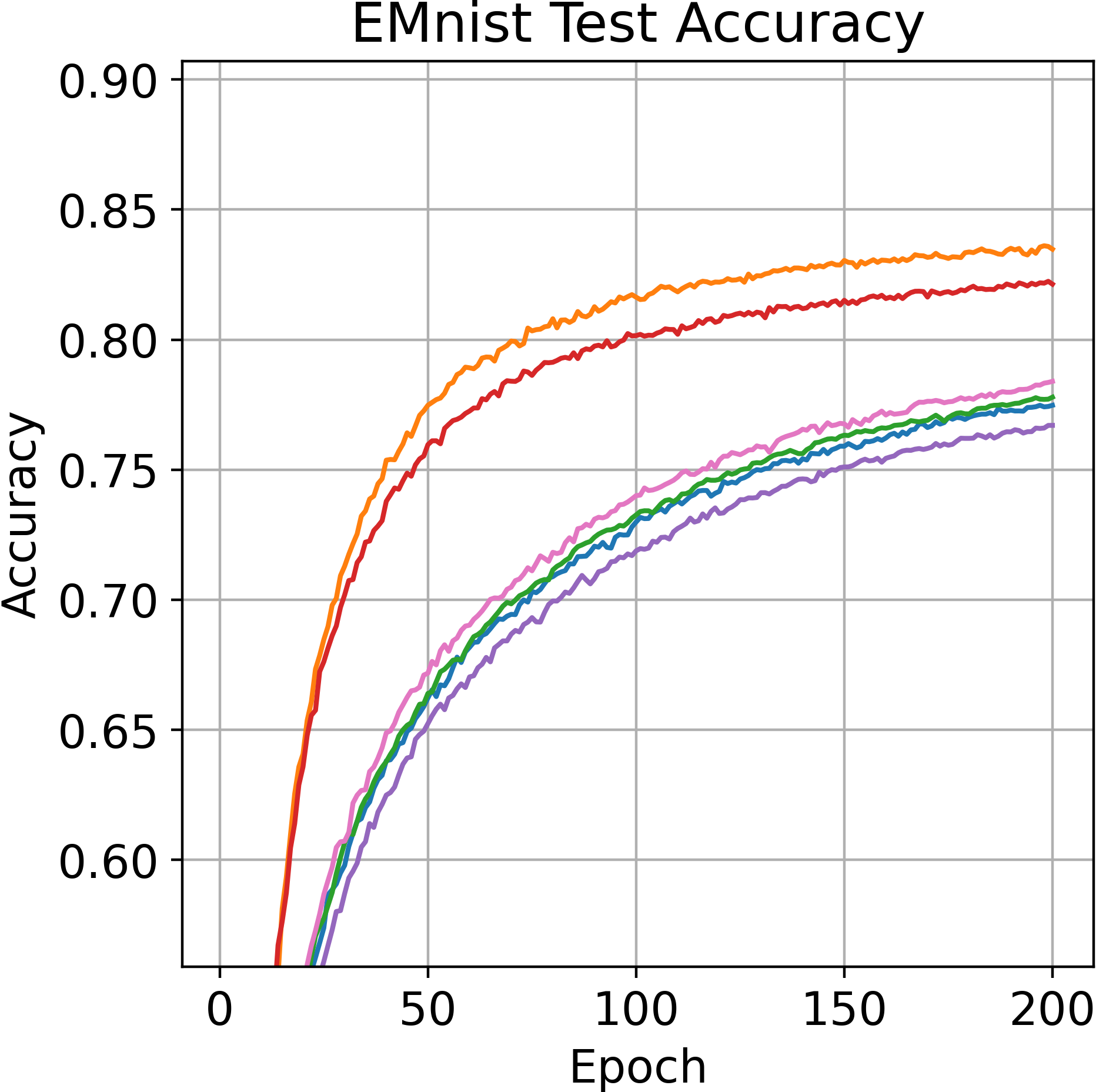}
	}
	\subfloat[$\alpha=10, T=20$]
	{
		\includegraphics[width=0.21\textwidth]{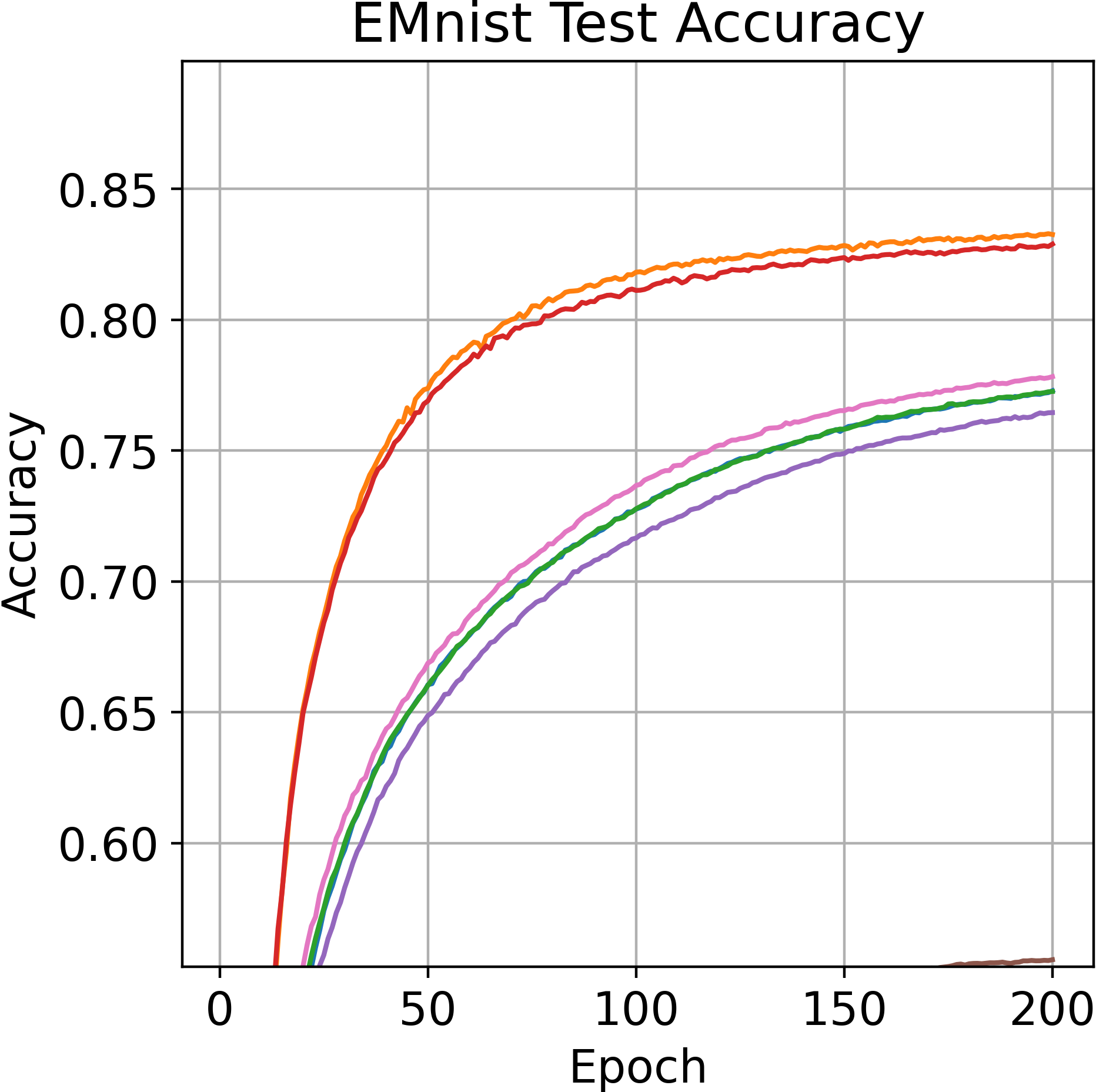}
	}
	\subfloat
	{
		\includegraphics[width=0.08\textwidth]{color-model.png}
	}

	\subfloat[$\alpha=0.05, T=40$]
	{
		\includegraphics[width=0.21\textwidth]{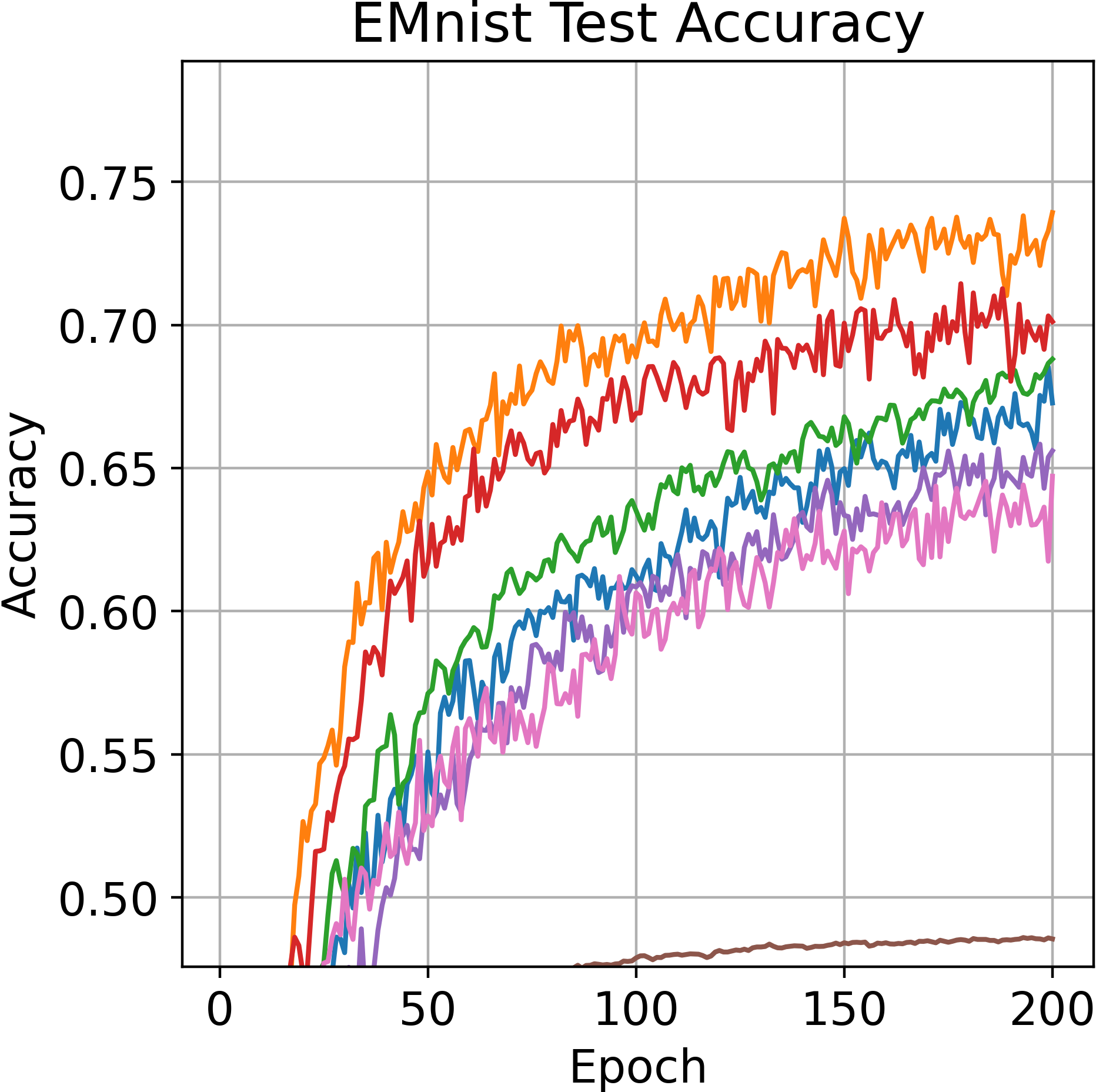}
	}
	\subfloat[$\alpha=0.1, T=40$]
	{
		\includegraphics[width=0.21\textwidth]{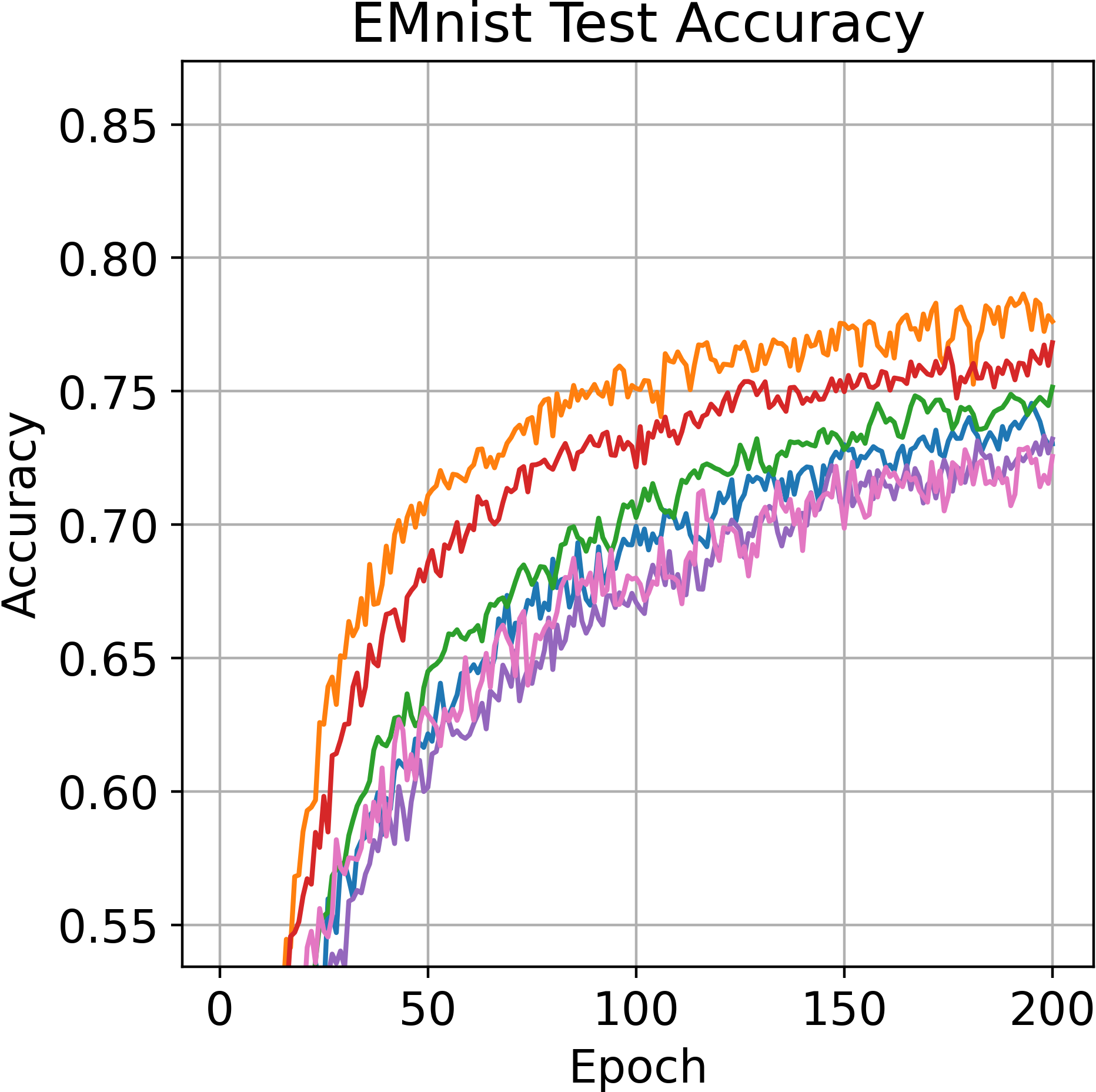}
  	}
	\subfloat[$\alpha=1, T=40$]
	{
		\includegraphics[width=0.21\textwidth]{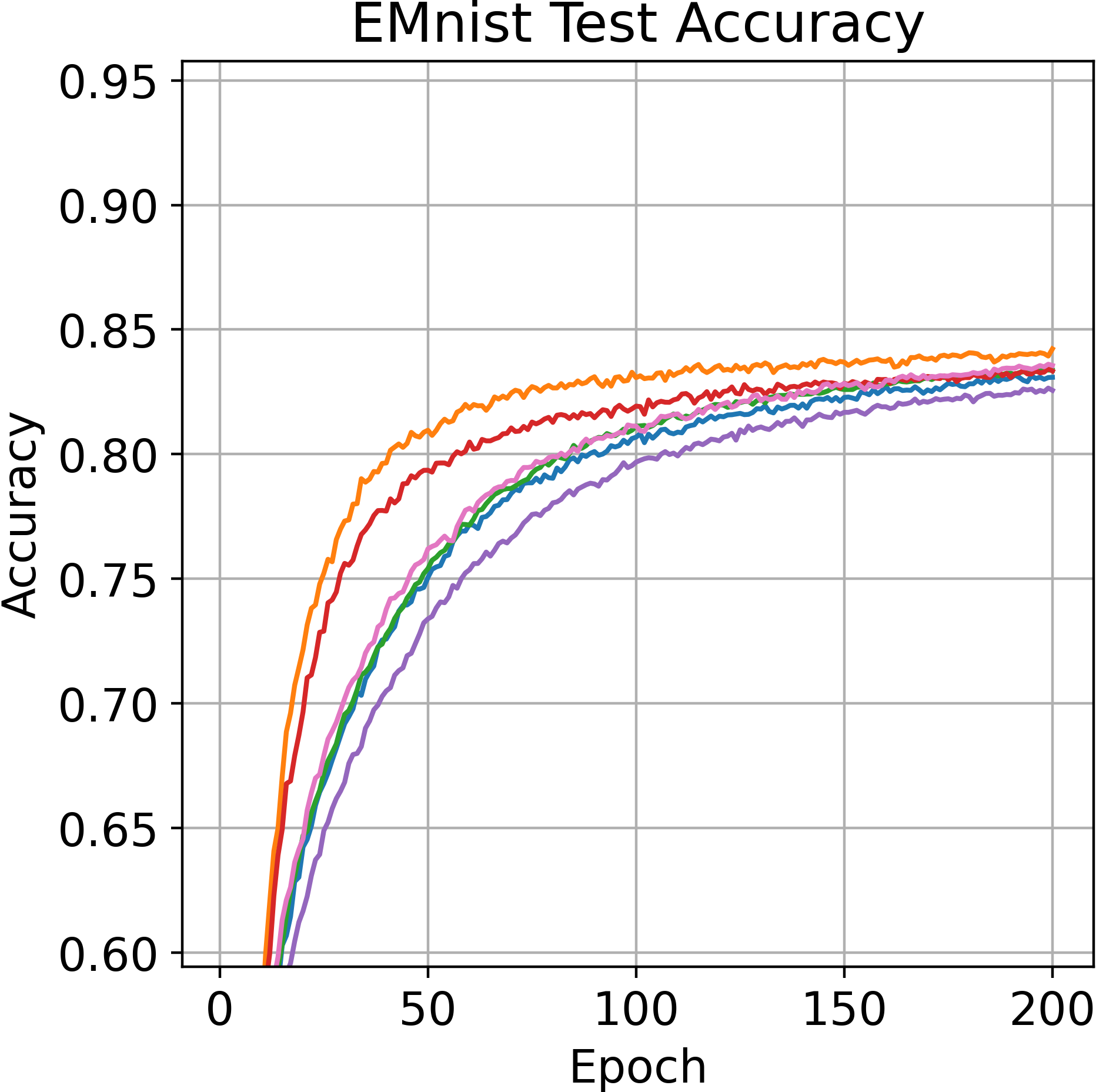}
	}
	\subfloat[$\alpha=10, T=40$]
	{
		\includegraphics[width=0.21\textwidth]{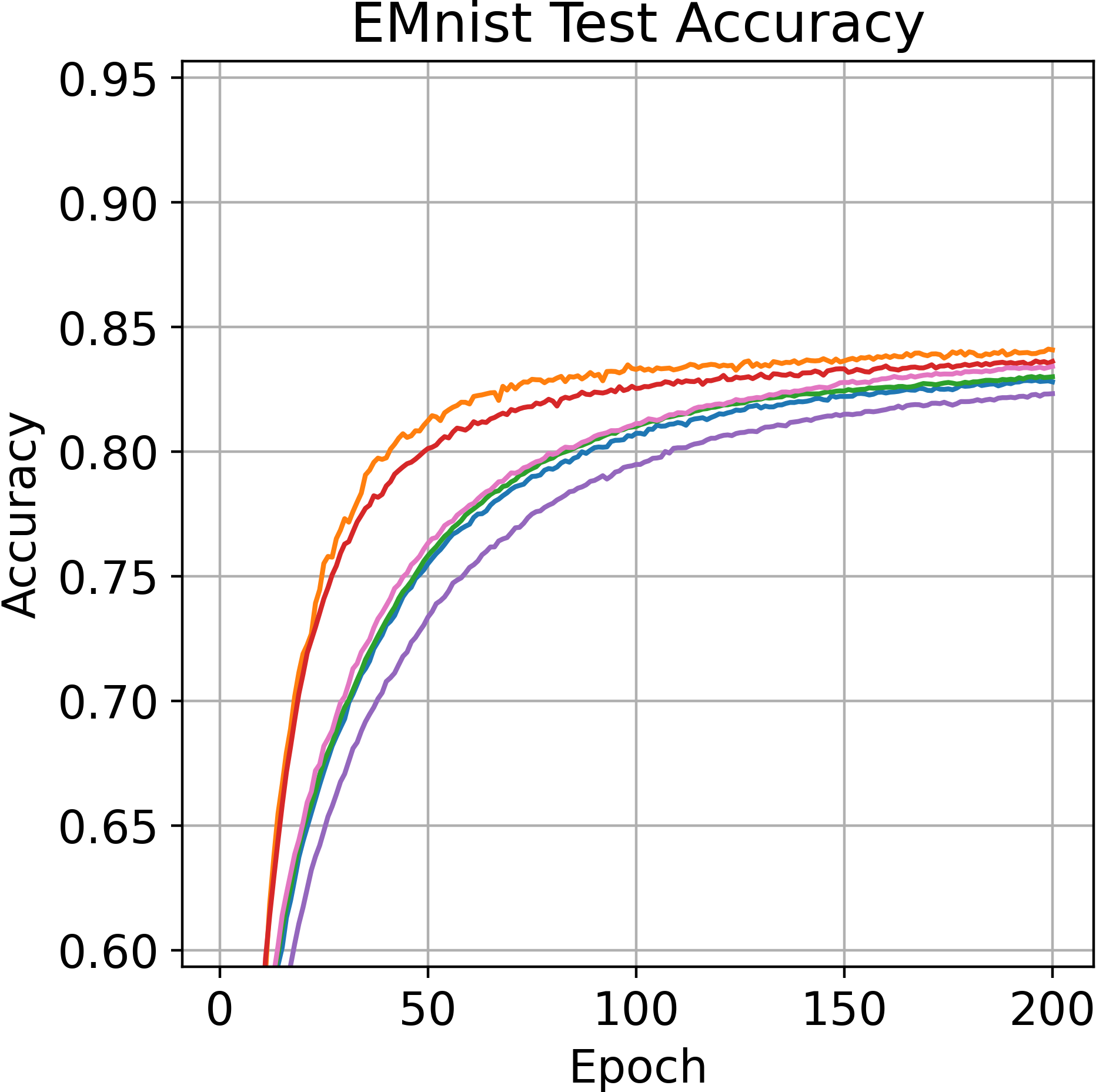}
	}
	\subfloat
	{
		\includegraphics[width=0.08\textwidth]{color-model.png}
	}

	\caption {Performance curves on EMNIST dataset, under different data heterogeneity and communication frequencies.}\label{ENMNIST Result}
\end{figure*}

\begin{figure*}[!h]
	\centering
	\subfloat
	{
			\includegraphics[width=0.21\textwidth]{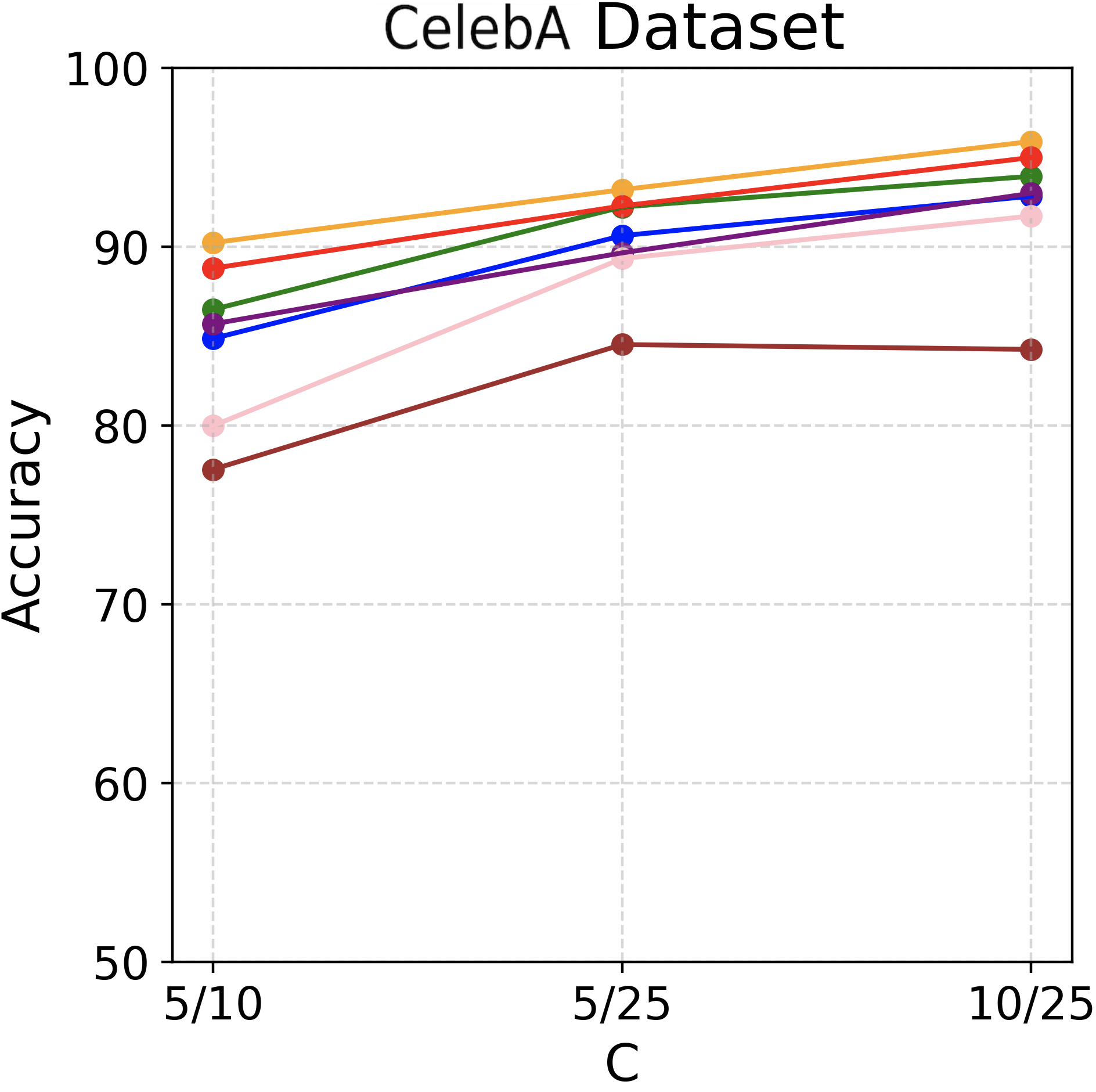}
			}
	\subfloat
	{
			\includegraphics[width=0.21\textwidth]{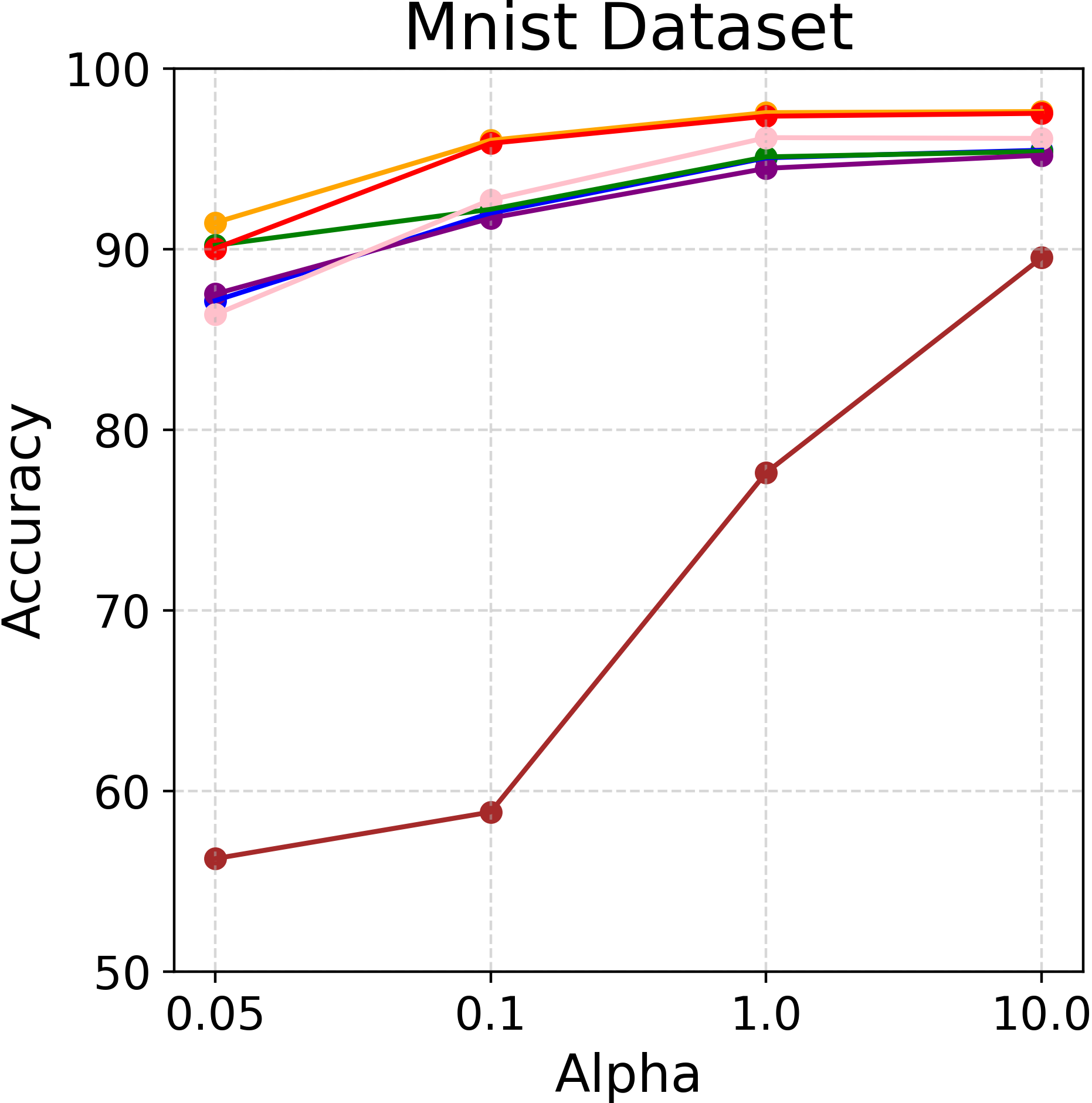}
	        }
    \subfloat
    {
	        \includegraphics[width=0.21\textwidth]{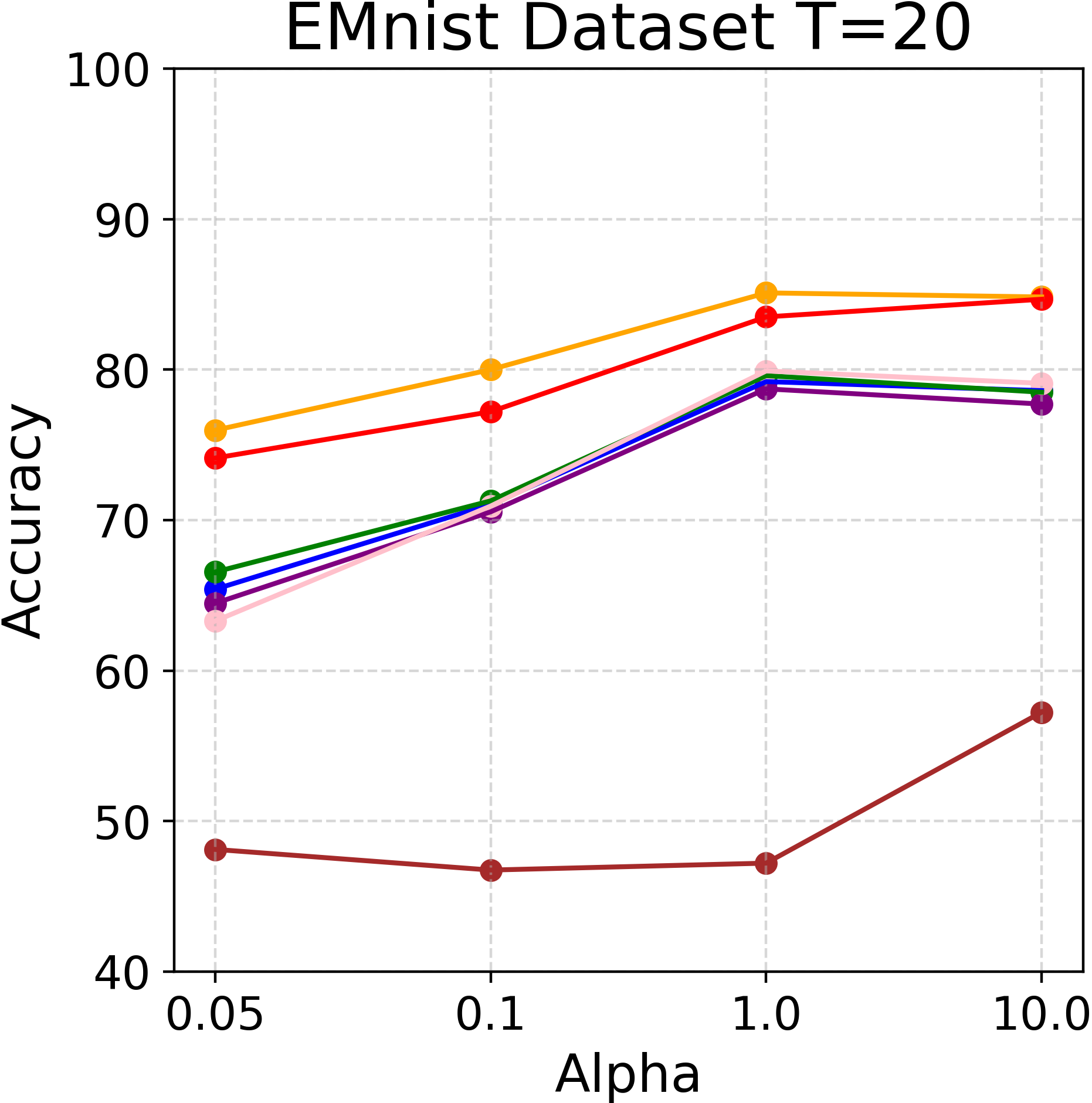}
            }
    \subfloat
    {
	        \includegraphics[width=0.21\textwidth]{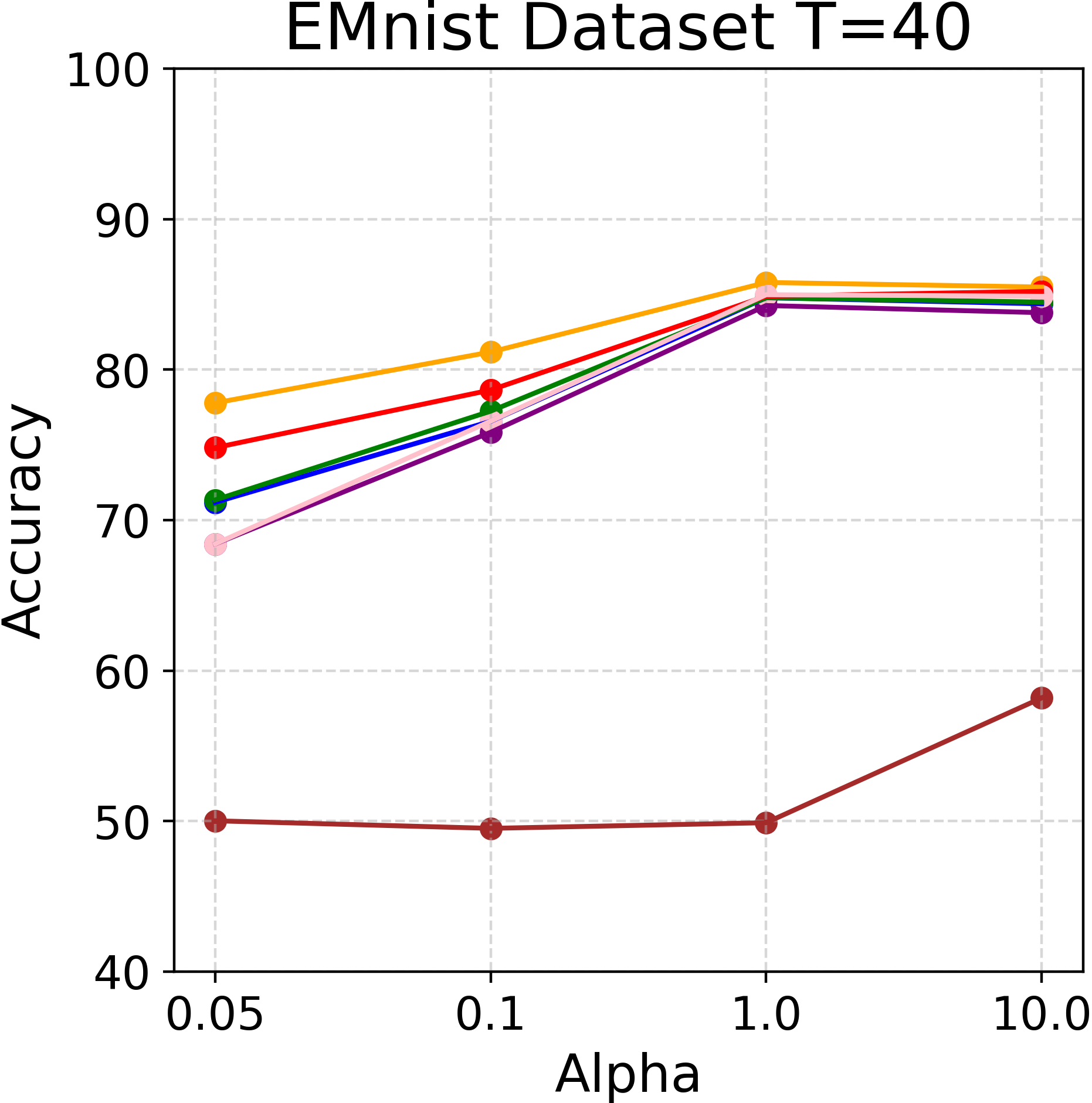}
            }
    \subfloat
    {
		    \includegraphics[width=0.08\textwidth]{color-model.png}
            }

	\caption {Visualized performance with respect to data heterogeneity, data distributions where smaller $\alpha$ indicates higher data heterogeneity.}
	\label{Performance Heterogeneity}
\end{figure*}

In this section, we empirically verify the effectiveness of FedCL. All experiments are conducted on a machine with an Intel Xeon CPU (4 GHz, 96 cores), 376 GB of DDR4 RAM, Nvidia Tesla V100-SXM2 with 32-GB HBM2 memory, and CUDA 11.6, running on Ubuntu 20.04.3. We summarize the setup of our experiment in Section \ref{sec: Experimental Setup}. Initially, FedCL is compared with six state-of-the-art approaches, including FedAvg \cite{mcmahan2017communication}, FedEnsemble \cite{shi2021fed}, FedGen \cite{zhu2021data}, FedProx \cite{li2020federated}, FedDistill \cite{jiang2020federated}, and FedDistill$^+$ \cite{zhu2021data}. Moreover, all experiments are conducted on three datasets, including MNIST \cite{deng2012mnist}, CELEBA \cite{liu2015deep}, and EMNIST \cite{cohen2017emnist}, with different setting parameters. In Section \ref{sec: Experimental Results}, we present the visualization of measurement in the training process, the experiment results, and their analysis. At the end of this section, ablation studies are conducted to verify the robustness of our algorithm. The source code can be found at \url{https://github.com/MingjieWang0606/FedCL_Pubic/}.

\subsection{Experimental Setup}
\label{sec: Experimental Setup}

\subsubsection{Datasets}

We use real-world datasets to comprehensively evaluate the performance of the proposed model. The datasets include MNIST \cite{deng2012mnist}, consisting of a training set of 60,000 examples and a test set of 10,000 examples; CELEBA \cite{liu2015deep}, comprising 202,599 face images with 40 attribute annotations; and EMNIST \cite{cohen2017emnist}, containing a training set of 697,932 examples and a test set of 116,323 examples. Following the suggestion of Zhu \textit{et al.} \cite{zhu2021data}, MNIST and EMNIST are both used for character and digit image classifications. On the other hand, CELEBA consists of celebrity faces, which is suitable for binary-classification tasks, such as predicting whether the celebrities in photos are smiling.

For heterogeneity exploration, we utilize the Dirichlet distribution \textbf{Dir}($\alpha$) to model Non-IID data distributions, where smaller $\alpha$ indicates higher data heterogeneity, making the distribution of $p_k(y)$ more biased for a user $k$ of MNIST and EMNIST, based on prior works \cite{lin2020ensemble}. In other words, a larger $\alpha$ results in greater differences in the distribution of labels for each client. Figure \ref{MNIST Result} shows the effect of adopting different $\alpha$ and different training baselines on the statistical heterogeneity for the MNIST dataset. For CELEBA, whose data is originally Non-IID, $r$ denotes the ratio of active users in total users, and $T$ is the local training step.

\subsubsection{Baselines}

We demonstrate the efficiency and effectiveness of FedCL by comparing it with other baselines. First, FedAvg \cite{mcmahan2017communication} is an aggregated model that iteratively averages the parameters of distributed local user models. Specifically, FedAvg randomly selects some users for sampling, replacing the non-sampled users with the current global model by averaging the gradient updates of these users to form a global update. FedEnsemble \cite{shi2021fed} is an extension of FedAvg that ensembles all user models' prediction output by using random permutations to update a group of models and then obtaining predictions through model ensembling by averaging. Moreover, FedGen \cite{zhu2021data} addresses heterogeneity in FL with knowledge distillation, and the distilled knowledge is directly adjusted to users' learning through distribution matching in the potential space. FedProx \cite{li2020federated} uses a proximal term in the objective function for stabilization, which improves robustness by regularizing local model training. Finally, FedDistill \cite{jiang2020federated} is a data-free knowledge distillation approach that shares the average of logit-vectors of labels among users and benefits from the robustness of the Bayesian model ensemble in aggregating users' predictions. However, the model parameters and its performance drop compared to other baselines. For comparison, FedDistill$^{+}$ \cite{zhu2021data} is used to share both network parameters and the logit-vectors of labels among users on average.

In our experiments, we evaluate the performance of the proposed FedCL and the aforementioned baselines using real-world datasets, including MNIST \cite{deng2012mnist}, CELEBA \cite{liu2015deep}, and EMNIST \cite{cohen2017emnist}. MNIST and EMNIST are used for character and digit image classifications, while CELEBA consists of celebrity faces and is used for a binary-classification task of predicting whether celebrities in photos are smiling. Besides, to ensure the validity and robustness of our experimental results, we employed 10 distinct random seeds in our study. We report the mean outcome of these 10 trials, along with their associated variance, as our final result.

\subsubsection{Training settings}
We train and test all datasets using the approaches with the same setting parameters. Hyperparameter configurations of the experiment are shown in Table \ref{Experiment Configurations}. For shared parameters of all models, 200 global communication rounds are performed except for CELEBA, and 20 user models are assigned, with the active-user ratio $r$ being 50\%. With a batch size of 32 and a local updating step $T$ of 20, we adopt the learning rate $r=0.01$ and use stochastic gradient descent (SGD) as the optimizer. For FedGen and FedCL, the optimizer for their generator is the Adam approach, and the learning rate is $10^{-4}$. Since FedDistill and FedDistill$^{+}$ both share the data of the logit-vector outputs of the user models for knowledge distillation, the distillation coefficient is 0.1. FedProx has a proximal term with a coefficient of 0.1 in the loss function. All of the above are implemented using PyTorch \cite{paszke2019pytorch}.

\subsection{Experimental Results}
\label{sec: Experimental Results}
\begin{table*}[!t]
	\caption{The number of rounds of different approaches to achieve the same accuracy as running FedAvg for 200 rounds (MNIST \& EMNIST) or 100 rounds (CELEBA).}
	\centering
        \scriptsize
		\setlength{\tabcolsep}{5.4mm}{
        \renewcommand\arraystretch{1.2}
			\begin{tabular}{l|c|c|c|c|c|c|c}
				\hline
				\multicolumn{8}{c}{\textbf{Top-1 Test Accuracy}}                                                                                                                   \\ \hline
				Dataset  & Approach & FEDAVG  & FEDPROX  & FEDENSEMBLE & FEDDISTILL & FEDGEN & FEDCL   \\ \cmidrule{1-8}
				\multirow{3}{*}{CELEBA}                                           & $r=5/10$  & 100 & 23 & 17 & $\backslash$ & 20 & \textbf{18} \\ \cmidrule{2-8}
				& $r=5/25$  & 100 & 58 & 16 & $\backslash$ & 15 & \textbf{17} \\ \cmidrule{2-8}
				& $r=10/2$5 & 100 & 37 & 22 & $\backslash$ & 19 & \textbf{16} \\ \cmidrule{1-8}
				\multirow{4}{*}{MNIST}                                            & $\alpha=0.05$  & 200 & 58 & 37 & $\backslash$ & 46 & \textbf{41} \\ \cmidrule{2-8}
				& $\alpha=0.1$   & 200 & 73 & 66 & $\backslash$ & 24 & \textbf{22} \\ \cmidrule{2-8}
				& $\alpha=1$     & 200 & $\backslash$ & 194 & $\backslash$ & 18 & \textbf{16} \\ \cmidrule{2-8}
				& $\alpha=10$    & 200 & $\backslash$ & $\backslash$ & $\backslash$ & 9 & \textbf{8} \\ \cmidrule{1-8}
				\multirow{4}{*}{\begin{tabular}[c]{@{}c@{}}EMNIST \\ $T=20$\end{tabular}} & $\alpha=0.05$  & 200 & 173 & 143 & $\backslash$ & 58 & \textbf{43} \\ \cmidrule{2-8}
				& $\alpha=0.1$   & 200 & $\backslash$ & 162 & $\backslash$ & 64 & \textbf{50} \\ \cmidrule{2-8}
				& $\alpha=1$     & 200 & $\backslash$ & 187 & $\backslash$ & 58 & \textbf{46} \\ \cmidrule{2-8}
				& $\alpha=10$    & 200 & $\backslash$ & $\backslash$ & $\backslash$ & 48 & \textbf{43} \\ \cmidrule{1-8}
				\multirow{4}{*}{\begin{tabular}[c]{@{}c@{}}EMNIST \\ $T=40$\end{tabular}} & $\alpha=0.05$  & 200 & $\backslash$ & 162 & $\backslash$ & 79 & \textbf{65} \\ \cmidrule{2-8}
				& $\alpha=0.1$   & 200 & 197 & 126 & $\backslash$ & 76 & \textbf{63} \\ \cmidrule{2-8}
				& $\alpha=1$     & 200 & $\backslash$ & 177 & $\backslash$ & 154 & \textbf{82} \\ \cmidrule{2-8}
				& $\alpha=10$    & 200 & $\backslash$ & 179 & $\backslash$ & 81 & \textbf{67} \\ \cmidrule{1-8}
		\end{tabular}
	}
	\label{Experiment Communication Rounds Results}
\end{table*}

\subsubsection{Accuracy and Analysis}

The data generator is a multilayer perceptron that takes a noise vector $\epsilon$ and a one-hot vector $y$ as inputs and generates a feature representation as output. The deep learning model is used as a classifier, with the last layer acting as the predictor \cite{mcmahan2017communication}. During training, half of the training dataset is distributed to users, while all testing datasets are reserved for performance evaluation. We also consider diversity loss to increase the diversity of the generator output. The learning curves for training on the MNIST, CELEBA, and EMNIST datasets are shown in Figures \ref{MNIST Result}, \ref{CELEBA Result}, and \ref{ENMNIST Result}.

Table \ref{Experiment Results} lists the top-1 test accuracy of all approaches with the default setting on MNIST, EMNIST, and CELEBA datasets. In these datasets, FedCL is the best performer and at least $0.2\%$ above the other baselines. The accuracy of FedGen is lower than that of FedCL, but it still yields better results than FedAvg.

\subsubsection{Data Heterogeneity}
FedCL and FedGen are two approaches that demonstrate robustness in the face of different levels of user heterogeneity while consistently performing well, as seen in Figure \ref{Performance Heterogeneity}. Specifically, Figure \ref{MNIST Result} shows that as the data distribution becomes more heterogeneous, the gain of FedCL is more notable with a smaller $\alpha$. FedCL performs the best because it takes into account the model-aware difficulty of the data, which helps to aggregate the models before model drifts occur. A key strength of both FedCL and FedGen is their ability to induce knowledge distilled to current users, reducing the difference in latent distributions across users. In contrast, FedAvg and FedProx cannot obtain shared information, and the curve of FedDistill drops dramatically due to knowledge distillation. These results highlight the importance of parameter sharing, which is vulnerable to data heterogeneities. Furthermore, the generator of FedCL is superior to that of FedGen, which may explain why FedEnsemble, which only ensembles predictions from user models, performs slightly lower than FedGen and FedCL. Overall, these findings demonstrate the effectiveness and efficiency of FedCL in addressing the challenges of user heterogeneity in FL.

\subsubsection{Learning Efficiency}
Regarding learning efficiency, FedCL has the fastest learning speed to reach optimized performance and outperforms other baselines. FedGen performs lower than FedCL, as each local user directly benefits from the learned knowledge. As shown in Table \ref{Experiment Communication Rounds Results}, compared with FedGen, FedCL reduces 4.3 rounds on CELEBA and 2.5 rounds on MNIST on average. The average reduction is 11.5 rounds on EMNIST with $T=20 $ and 28.25 rounds on EMNIST with $T=40 $. Our model is robust against different levels of communication delays, but the accuracy advantage of FedCL over FedGen varies among datasets, depending on the dataset distribution. For other approaches, their trends consistently underperform FedCL by a significant margin, which becomes more pronounced with high data heterogeneity. Therefore, FedCL not only achieves fast convergence but also has less communication workload when given a wise parameter sharing strategy. Our approach can directly benefit every local user with learned knowledge, being more explicit and consistent.

\begin{table}[!t]
	\caption{Performance Overview under Different Threshold for Moving to Next Difficulty Level}
	\centering
        \scriptsize
		\setlength{\tabcolsep}{3mm}{
        \renewcommand\arraystretch{1.2}
			\begin{tabular}{c|cccc}
				\hline
				\multicolumn{5}{c}{\textbf{Performance w.r.t. different P-Threshol}}\\
				\multicolumn{5}{c}{\textbf{on EMNIST $T=20$ $\alpha=0.1$}} \\ \cmidrule{1-5}
				\textbf{P-Value} & 60 & 70 & 80 & 90 \\ \cmidrule{1-5}
				\textbf{Accuracy} & \multicolumn{4}{c}{FedCL=78.06±1.25}\\ \cmidrule{1-5}
				FEDCL & 78.02±0.75 & 77.68±0.85 & 78.06±1.25 & 77.55±0.61 \\ \cmidrule{1-5}
		\end{tabular}
	}
	\label{p-threshold experiment}
\end{table}

\begin{table}[!t]
	\caption{Performance Overview under Different Difficulty Level Division}
	\centering
		\scriptsize
		\setlength{\tabcolsep}{3mm}{
        \renewcommand\arraystretch{1.2}
			\begin{tabular}{c|cccc}
				\hline
				\multicolumn{5}{c}{\textbf{Performance w.r.t. different Stage-num}}                                                                            \\
				\multicolumn{5}{c}{\textbf{on EMNIST $T=20$ $\alpha=0.1$}}                                                                                     \\ \cmidrule{1-5}
				\textbf{Stage-num} & 2          & 3          & 4          & 5          \\ \cmidrule{1-5}
				\textbf{Accuracy}                          & \multicolumn{4}{c}{FedCL=78.06±1.25}                                     \\ \cmidrule{1-5}
				FEDCL              & 78.05±1.22 & 78.01±0.73 & 78.06±1.25 & 77.61±0.93 \\ \cmidrule{1-5}
	\end{tabular}
	}
	\label{Muti-stage Experiment}
\end{table}

\subsubsection{Hyperparameter Experiments}
\begin{table}[!t]
        \scriptsize
		\setlength{\tabcolsep}{1mm}{
        \renewcommand\arraystretch{1.2}
	\caption{Performance w.r.t Different Hyperparameters.}
	\centering
			\begin{tabular}{c|cccc}
				\cmidrule{1-5}
                \multicolumn{5}{c}{\textbf{Performance Overview under Different  Hyperparameters}}
                \\
                \multicolumn{5}{c}{\textbf{on EMNIST $T=20$ $\alpha=0.1$}}\\\hline
                $\lambda$&$0.2$&$0.5$&$0.8$&$1.0$ \\ \cmidrule{2-5}
                $\tau=10$&77.29$\pm0.43$&78.06$\pm1.25$&77.83$\pm0.23$&78.01$\pm0.81$ \\ \cmidrule{1-5}
                 
                $\tau$&$5$&$10$&$15$&$20$ \\ \cmidrule{2-5}
                $\lambda=0.5$&77.96$\pm1.02$&78.06$\pm1.25$&78.02$\pm0.52$&77.84$\pm0.64$ \\ \cmidrule{1-5}
                
                $\mathcal{S}$&\{0.2,0.4,0.6,0.8\}&\{0.3,0.5,0.7\}&\{0.2,0.6,0.8\}&\{0.4,0.8\} \\ \cmidrule{2-5}
                $\lambda=0.5,\tau=10$&77.56$\pm0.15$&77.86 $\pm0.23$ &78.06 $\pm1.25$ & 77.31 $\pm0.54$ \\ 
                \cmidrule{1-5}

                $v$&0.5&0.6&0.7&0.8 \\ \cmidrule{2-5}
                $\lambda=0.5,\tau=10$&76.01$\pm0.52$&77.31$\pm0.71$&78.06$\pm1.25$&77.51$\pm0.82$ \\ 
                \cmidrule{1-5}
                
		\end{tabular}
	}
	\label{HE}
\end{table}

We conducted hyperparameter experiments to investigate the impact of different hyperparameters on the performance of our proposed method. These experiments were carried out on the EMNIST dataset with $T=20$ and $\alpha=0.1$. In the first set of hyperparameter experiments, we {varied} $\lambda$, the hyperparameter of the regularization term $\lambda(\log\sigma_i)^2$ in $l^{cl}$, while keeping $\tau=10$ fixed. Our results {indicated} that the best performance was achieved with $\lambda=0.5$. In the second set of hyperparameter experiments, we varied $\tau$, the loss-based threshold that distinguishes easy and hard samples, while keeping $\lambda=0.5$ fixed. Our results revealed that the best performance was achieved with $\tau=10$.

In the third and fourth experiments, we varied the array of training states $\mathcal{S}$ and the temperature coefficient $v$ of $N_k$ to represent the number of samples. These experiments {were conducted} with the best $\tau$ and $\lambda$ values obtained from the first two groups of experiments. Our results demonstrated that the best performance was achieved when $\mathcal{S}={0.3,0.5,0.7}$ and $v=0.6$. Overall, our hyperparameter experiments showed that the proposed method is insensitive to the choice of hyperparameters.

\subsubsection{Ablation Experiments}
Extended analysis indicated that FedCL is robust across different thresholds for curriculum satisfaction. As shown in Table \ref{Muti-stage Experiment}, our default threshold is 80\%, which means when over 80 percent of clients {have finished} over 80 percent of the corresponding curriculum data, the training process will move to the next difficulty level. In Table \ref{p-threshold experiment}, the performance of FedCL was measured under different thresholds of 60, 70, 80, and 90, and it outperformed FedGen for all four situations by at least 1.5\% accuracy.

Furthermore, we explored different numbers of difficulty level settings in Table \ref{Muti-stage Experiment}. In Table \ref{p-threshold experiment}, there are three steps set as default. For example, the difficulty level will move from 0.3 to 0.6 and 0.9. For other numbers of steps, it can be deduced by analogy. For step numbers shown in Table \ref{p-threshold experiment}, FedCL outperformed FedGen overall by at least 1.5\% accuracy. Therefore, compared to FedGen, FedCL has a stronger generalization ability.

\section{Conclusion and Future work}

Our paper introduces FedCL, a practical and effective approach for enhancing the performance of federated deep learning models on Non-IID datasets. The proposed approach employs synchronous FL, a new learning concept that leverages the learning progress of the model to address heterogeneity issues and mitigate model drift. Through extensive experiments on four benchmarks, the authors demonstrate that FedCL outperforms state-of-the-art methods, achieving better generalization performance with fewer communication rounds. The authors propose that future research could focus on the trade-off analysis between privacy and utility based on synchronous FL and the integration of the learning progress of the model for guaranteed data privacy. We conclude that FedCL is a promising FL method that can effectively tackle challenges in distributed learning scenarios and improve the performance of FL systems in real-world applications.

\section*{Acknowledgements}
This work was supported in part by the Guangdong Key Lab of AI and Multi-modal Data Processing, BNU-HKBU United International College (UIC) under Grant No. 2020KSYS007 and Computer Science Grant No. UICR0400025-21; the National Natural Science Foundation of China (NSFC) under Grant No. 61872239 and No. 62202055; the Institute of Artificial Intelligence and Future Networks, Beijing Normal University; the Zhuhai Science-Tech Innovation Bureau under Grants No. ZH22017001210119PWC and No. 28712217900001; and the Interdisciplinary Intelligence Supercomputer Center of Beijing Normal University (Zhuhai).

\ifCLASSOPTIONcaptionsoff
  \newpage
\fi

\bibliographystyle{IEEEtran}
\bibliography{references}

\begin{IEEEbiography}[{\includegraphics[width=1in,height=1.25in,clip,keepaspectratio]{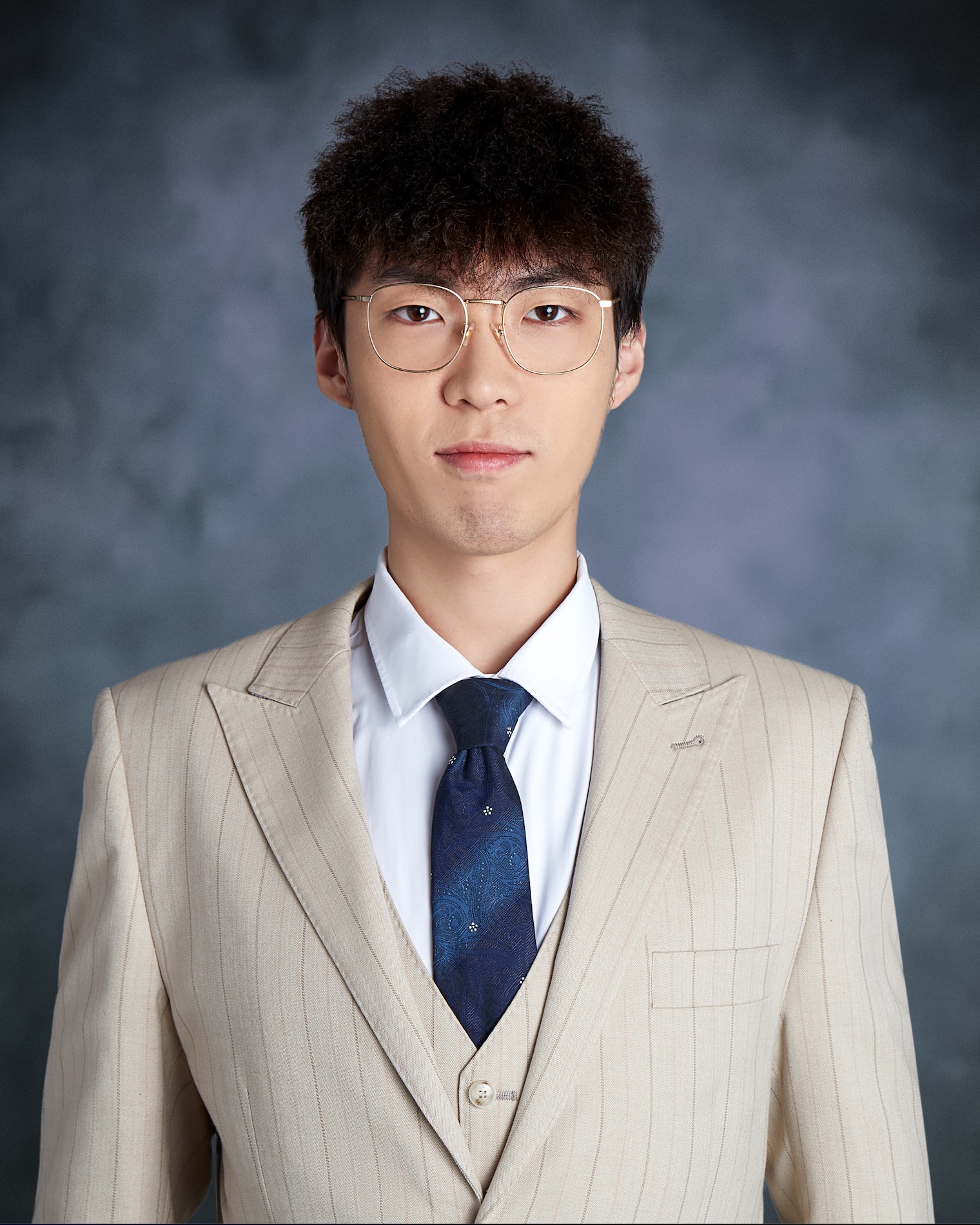}}]{Mingjie Wang}
    received the B.S. degree from Department of Computer Science and Technology, Longdong University, China, in 2021 and is currently a M.Phil. candidate in Department of Data Science and Technology, BNU-HKBU United International College (UIC). His current research interests include Time-Series, NLP, Machine Learning and Deep Learning.
\end{IEEEbiography}

\begin{IEEEbiography}[{\includegraphics[width=1in,height=1.25in,clip,keepaspectratio]{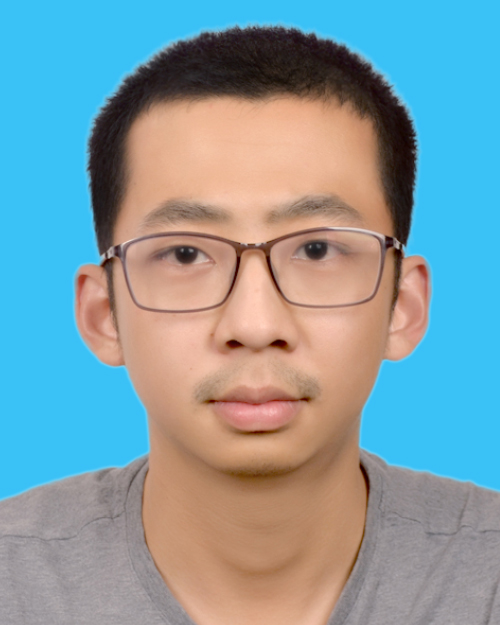}}]{Jianxiong Guo}
	received his Ph.D. degree from the Department of Computer Science, University of Texas at Dallas, Richardson, TX, USA, in 2021, and his B.E. degree from the School of Chemistry and Chemical Engineering, South China University of Technology, Guangzhou, China, in 2015. He is currently an Assistant Professor with the Advanced Institute of Natural Sciences, Beijing Normal University, and also with the Guangdong Key Lab of AI and Multi-Modal Data Processing, BNU-HKBU United International College, Zhuhai, China. He is a member of IEEE/ACM/CCF. He has published more than 40 peer-reviewed papers and been the reviewer for many famous international journals/conferences. His research interests include social networks, wireless sensor networks, combinatorial optimization, and machine learning.
\end{IEEEbiography}

\begin{IEEEbiography}[{\includegraphics[width=1in,height=1.25in,clip,keepaspectratio]{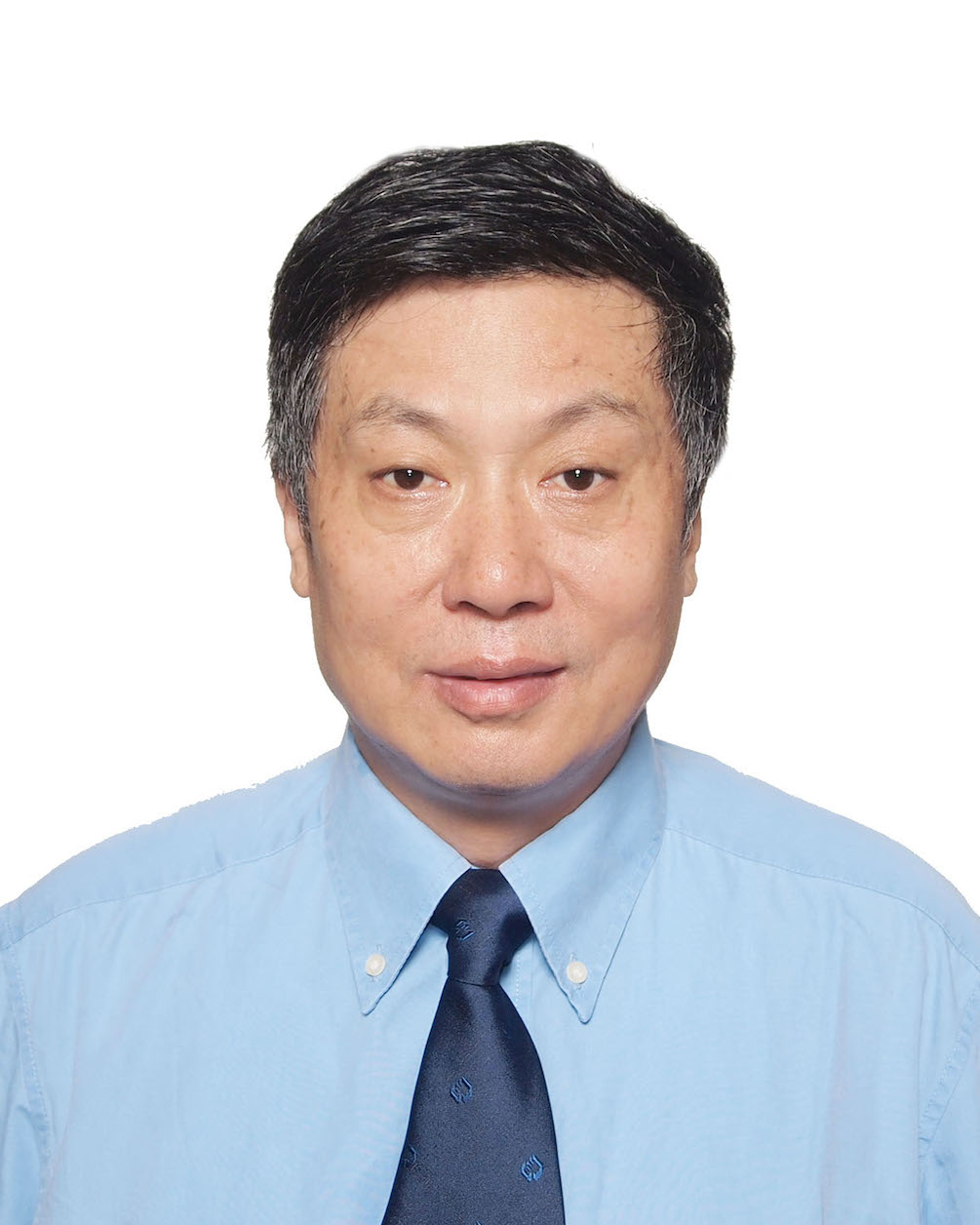}}]{Weijia Jia}
	is currently a Chair Professor, Director of BNU-UIC Institute of Artificial Intelligence and Future Networks, Beijing Normal University (Zhuhai) and VP for Research of BNU-HKBU United International College (UIC) and has been the Zhiyuan Chair Professor of Shanghai Jiao Tong University, China. He was the Chair Professor and the Deputy Director of State Kay Laboratory of Internet of Things for Smart City at the University of Macau. He received BSc/MSc from Center South University, China in 82/84 and Master of Applied Sci./PhD from Polytechnic Faculty of Mons, Belgium in 92/93, respectively, all in computer science. For 93-95, he joined German National Research Center for Information Science (GMD) in Bonn (St. Augustine) as a research fellow. From 95-13, he worked in City University of Hong Kong as a professor. His contributions have been recognized as optimal network routing and deployment; anycast and QoS routing, sensors networking, AI (knowledge relation extractions; NLP etc.) and edge computing. He has over 600 publications in the prestige international journalsconferences and research books and book chapters. He has received the best product awards from the International Science \& Tech. Expo (Shenzhen) in 2011/2012 and the 1st Prize of Scientific Research Awards from the Ministry of Education of China in 2017 (list 2). He has served as area editor for various prestige international journals, chair and PC memberskeynote speaker for many top international conferences. He is the Fellow of IEEE and the Distinguished Member of CCF.
\end{IEEEbiography}

\end{document}